\PassOptionsToPackage{table}{xcolor}
\documentclass{article} 
\usepackage{iclr2027_conference,times}
\usepackage[T1]{fontenc}

\usepackage{amsmath,amsfonts,bm}

\def\eqref#1{equation~\ref{#1}}

\def\1{\bm{1}}

\DeclareMathAlphabet{\mathsfit}{\encodingdefault}{\sfdefault}{m}{sl}
\SetMathAlphabet{\mathsfit}{bold}{\encodingdefault}{\sfdefault}{bx}{n}

\usepackage{hyperref}
\usepackage{url}
\usepackage{amsmath,amssymb,amsthm}
\usepackage{booktabs}
\usepackage{array}
\usepackage{enumitem}
\usepackage{graphicx}
\usepackage[table]{xcolor}
\usepackage{tikz}
\usetikzlibrary{arrows.meta,positioning,fit,backgrounds,calc}
\usepackage{microtype}
\usepackage[nameinlink]{cleveref}
\crefname{figure}{Fig.}{Figs.}
\Crefname{figure}{Figure}{Figures}
\crefname{table}{Tab.}{Tabs.}
\Crefname{table}{Table}{Tables}
\crefname{section}{Sec.}{Secs.}
\Crefname{section}{Section}{Sections}
\crefname{appendix}{App.}{Apps.}
\Crefname{appendix}{Appendix}{Appendices}
\crefname{equation}{Eq.}{Eqs.}
\Crefname{equation}{Equation}{Equations}
\crefname{proposition}{Prop.}{Props.}
\Crefname{proposition}{Proposition}{Propositions}
\crefname{assumption}{Assum.}{Assums.}
\Crefname{assumption}{Assumption}{Assumptions}
\crefname{definition}{Def.}{Defs.}
\Crefname{definition}{Definition}{Definitions}
\usepackage{subcaption}

\usepackage{wrapfig}

\makeatletter
\AtBeginDocument{\let\READorigcaption\caption}

\makeatother

\theoremstyle{definition}

\title{READ: Reading the Selective Scan for Faithful Video Attribution}

\author{%
Magdalena Trędowicz$^{1}$, Piotr Kubaty$^{1}$, Arkadiusz Lewicki$^{2}$,
\textbf{Jacek Tabor$^{1,3}$, Łukasz Struski$^{1}$} \\[4pt]
$^{1}$Jagiellonian University, Kraków, Poland \\
$^{2}$University of Information Technology and Management in Rzeszów, Poland \\
$^{3}$Centre for Credible AI, Warsaw University of Technology, Warsaw, Poland \\[2pt]
\texttt{\{magdalena.tredowicz,piotr.kubaty\}@doctoral.uj.edu.pl}
}

\newcommand{\our}{READ}

\iclrfinalcopy 
\begin{document}

\maketitle


\vspace{-1.6em}
\begin{center}
    \includegraphics[
        width=\textwidth,
        trim=0.5cm 0 0 16.5,
        clip
    ]{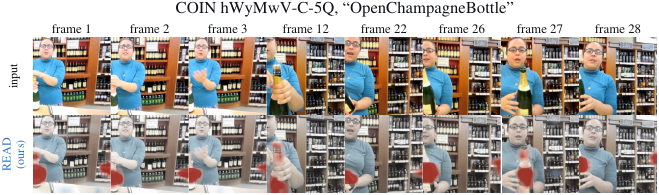}
\captionof{figure}{\textbf{\our{} localises the object relevant to the procedure.} Eight frames from a COIN clip of \emph{Open a champagne bottle}, selected by temporal relevance under \our{} (top: input; bottom: \our{} attribution). Attribution consistently concentrates on the bottle across the procedure. Red indicates positive attribution. The map is obtained from a single forward pass by decomposing the scan readout into per-token contributions.}
    \label{fig:teaser}
\end{center}
\vspace{-0.4em}

\begin{abstract}
Selective state-space models are an efficient approach to long-sequence video modeling. However, their recurrent computation makes it difficult to determine which video tokens drive a prediction. Existing attribution methods often require additional backward passes or repeated model evaluations, making them substantially more expensive than the forward inference they aim to explain. In this work, we introduce \our{}, a parameter-free attribution method for selective state-space models that unrolls the recurrence at the position read by the classifier. \our{} traces the contribution accumulated in the recurrent state back to the video tokens that contribute to the final prediction. By anchoring attribution at the classifier's readout, it produces a single spatiotemporal relevance map from which frame- and region-level relevance can be obtained directly. Despite requiring only one forward pass, \our{} achieves competitive faithfulness and localisation compared with substantially more expensive attribution methods. Experimental results show that \our{} avoids the systematic relevance biases of existing methods while recovering meaningful temporal and class-specific structure across datasets and backbones. More broadly, our results show that the recurrence underlying selective state-space models is not only an efficient mechanism for inference, but also a direct source of interpretable evidence for their predictions.

\end{abstract}

\section{Introduction}

Sequence modelling is a fundamental problem across language, speech, time series, and vision. For many years, recurrent and convolutional architectures provided the main tools for neural sequence modelling. The Transformer then introduced an alternative based on self-attention, which models content-dependent interactions between tokens~\cite{vaswani2017attention}. However, the quadratic computational cost of standard self-attention becomes restrictive for long sequences~\cite{tay2020efficient}. State-space models (SSMs) can instead process sequences in linear time by updating a fixed-size latent state at each step. S4 demonstrated efficient modelling of long-range dependencies~\cite{gu2022efficiently}. Mamba introduced selective SSMs with input-dependent dynamics that control how information is retained or discarded, supporting applications to language, audio, and genomics~\cite{gu2024mamba}. Subsequent work adapted this approach to images~\cite{hatamizadeh2025mambavision,liu2024vmamba}.

Video provides a particularly natural setting for SSMs, as the number of tokens grows with both the number of frames and spatial resolution. This has motivated architectures such as VideoMamba, which applies selective SSMs to video understanding~\cite{li2024videomamba}.
A key difference from Transformers lies in how token interactions are represented. Transformer attention matrices expose these interactions explicitly, providing a starting point for attribution, although attention weights alone do not necessarily yield faithful explanations~\cite{jain2019attention,jacovi2020attention}.
In selective SSMs, information is accumulated through recurrent state updates, with token interactions remaining implicit in this computation. Interpreting their predictions therefore involves tracing the accumulated evidence back to the regions and frames that contributed to it.

MambaLRP addresses this challenge by adapting relevance propagation to selective dynamics and multiplicative gates~\cite{rezaei2024mambalrp}. However, backward-based attribution requires storing activations across layers, increasing memory use for long video sequences. The scan order and readout position can also affect the distribution of temporal relevance. In VideoMamba's backward scan, tokens from the first frame immediately precede the classifier's readout position. These considerations raise a practical question: can the information accumulated at the classifier's readout be traced directly to video tokens, providing faithful spatial and temporal explanations without backward propagation?

\begin{figure}[t]
    \centering
    \includegraphics[width=\linewidth]{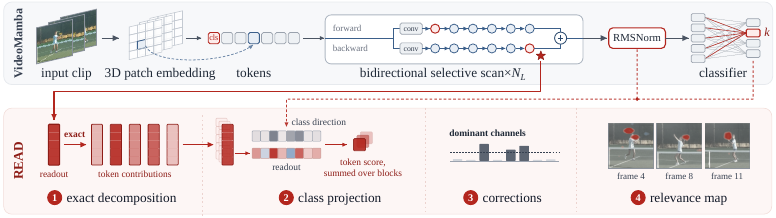}
    \caption{\textbf{Overview of \our{}.} Top: the frozen VideoMamba backbone. The class token (\texttt{cls}) closes the backward scan, making its final state the classifier readout. Bottom: \our{} decomposes this readout in a single forward pass. Unrolling the selective scan gives an exact decomposition into per-token contributions. Each contribution is projected onto the target class through the readout and the block-wise class direction. The resulting scores are aggregated across blocks. We correct the systematic offset introduced by the causal convolution, account for dominant channel contributions, and suppress cancellations across blocks. The resulting signed scores form a single spatiotemporal relevance map and temporal profile. Red denotes evidence for the target class, blue evidence against it. Example shows frames of a Kinetics-400 clip.}
    \label{fig:architecture}
    \vspace{-0.5cm}
\end{figure}

We introduce \our{}, a parameter-free attribution method that traces information accumulated in a selective scan back to individual video tokens.
Once the input-dependent coefficients are fixed, the readout state is a linear function of the scan inputs and can therefore be decomposed exactly into per-token contributions.
We use them to estimate class-specific relevance, accounting for the final normalisation and correcting the spatial offset introduced by the preceding causal convolution.
\our{} operates on the frozen model in a single forward pass, producing a spatiotemporal map that provides spatial and temporal views of the evidence supporting the prediction. For the architecture overview see~\cref{fig:architecture}. Our main contributions are as follows:
\begin{itemize}[leftmargin=1.2em,itemsep=2pt,topsep=2pt]
\item We introduce \our{}, a parameter-free attribution method that decomposes a selective scan's readout state into exact per-token contributions and uses them to construct a class-specific spatiotemporal map, providing spatial and temporal views of the evidence supporting a prediction.

\item We derive a class-sensitive readout projection and a convolution-offset correction that map recurrent-state contributions back to the input tokens. The projection uses a path-averaged readout direction, while the correction aligns contributions with the tokens that generated them.

\item We develop a streaming implementation that accumulates attribution contributions block by block in a single forward pass, without backward propagation. Its auxiliary memory is independent of model depth, substantially reducing memory requirements relative to the evaluated backward-based baselines.

\item We evaluate \our{} across three video datasets and three backbone sizes, showing consistent performance across faithfulness and localisation benchmarks, with sanity checks validating the resulting attributions.

\end{itemize}


\section{Related Work}
\label{sec:related}

\paragraph{Post-hoc attribution methods.}\hspace{-0.3cm}
Post-hoc attribution methods relate predictions to input features through different signals. Grad-CAM and Layer-CAM combine intermediate activations with gradients to produce class-specific maps~\citep{selvaraju2017gradcam,jiang2021layercam}, while Integrated
Gradients accumulates gradients along a path from a reference input to the observed sample~\citep{sundararajan2017axiomatic}.
Perturbation-based methods instead assess prediction changes under modified inputs, with RISE estimating relevance through randomized masks~\citep{petsiuk2018rise}.
These methods provide broadly applicable baselines, although backward computation
and repeated evaluations can make attribution costly for long video sequences.
A complementary direction develops attribution rules around the mechanisms through
which a particular architecture transforms and accumulates information.

\vspace{-0.3cm}\paragraph{Attribution for recurrent and selective state-space models.}\hspace{-0.3cm}
For recurrent architectures, early work adapted relevance propagation to the
multiplicative interactions of LSTMs~\citep{arras2017explaining}, while REAT expresses
predictions of gated RNNs as additive token contributions~\citep{du2019attribution}.
These formulations do not transfer directly to Mamba, whose input-dependent dynamics
govern information injection, propagation, and readout~\citep{gu2024mamba}.
MambaAttr represents the selective scan through implicit token interactions and
combines them with gradients to obtain class-specific maps~\citep{ali2025hidden}.
Explicitly materialising this interaction operator incurs quadratic storage cost
in sequence length (\cref{app:baseline_choice}).
MambaLRP instead propagates relevance backward using conservation-based rules
tailored to selective dynamics, nonlinearities, and gates~\citep{rezaei2024mambalrp}.
LaTIM offers a complementary forward decomposition of token-to-token contributions
in Mamba, with its ALTI-Logit variant formulated for output-token logits in language
models~\citep{pitorro2025latim}.
For video classification, the target shifts to a clip-level prediction, and the
challenge is to relate its evidence to the spatial and temporal locations of visual tokens.

\vspace{-0.3cm}\paragraph{Spatiotemporal explanations for video models.}\hspace{-0.3cm}
The importance of this temporal structure is reflected in studies showing both the
role of temporal ordering in distinguishing actions and the tendency of learned
representations to favor static information~\citep{sevilla2021only,kowal2022deeper}.
Attention weights provide one route to examining the distribution of evidence:
attention rollout aggregates them across Transformer layers to estimate input-token
relevance~\citep{abnar2020quantifying}, while STAA derives spatial and temporal
importance in video Transformers with a single forward pass~\citep{wang2025staa}.
Selective video SSMs such as VideoMamba instead accumulate information through
recurrent scans, where scan direction and readout position shape how visual tokens
contribute to a prediction~\citep{li2024videomamba}.
The preceding causal convolution also mixes neighboring tokens, so contributions
associated with a scan position may include information from nearby patches.
Explaining these models therefore involves tracing evidence accumulated at the
classifier's readout back to the contributing regions and frames, while accounting
for local mixing of information.
\vspace{-0.3cm}
\section{\our{}: Readout-Anchored Decomposition}
\label{sec:method}

\vspace{-0.3cm}

Our goal is to develop a post-hoc attribution method that identifies the spatiotemporal tokens contributing to a prediction of a selective SSM. A selective SSM does not expose an interaction matrix that directly reveals token contributions. Instead, each input token contributes to the recurrent state, with its contribution transformed by the subsequent scan steps between that token and the position read by the classifier.

The classifier reads the state at one position, which we call the \emph{readout}. For a fixed input clip, the input-dependent coefficients of the selective scan are fixed, leaving a linear recurrence. The state at the readout can therefore be written exactly as a sum of token-wise contributions, with one contribution for each input token. Unrolling the recurrence gives this decomposition in closed form from a single forward pass, without gradients, surrogate operators, or repeated model evaluations. The decomposition itself is exact; mapping the token-wise state contributions to the target logit requires a first-order approximation.

Three design choices determine how this decomposition is converted into an attribution map.

\emph{In which direction is the sum read?} A token contribution is a vector, whereas the prediction is a scalar. A natural choice is the classifier direction, but the classifier operates on the residual stream through a normalisation, under which a contribution that only changes the stream magnitude may have little or no effect on the logit. We therefore project onto a logit-sensitive direction obtained by averaging the local sensitivity along the path through which a block transforms the residual stream (\cref{sec:readout_direction}).

\emph{Where does the credit belong?} A causal depthwise convolution precedes the scan, so the input at a scan position mixes the corresponding token with its predecessors. Without correction, this shifts the resulting attribution in time. We therefore redistribute each block's credit along the taps of the convolution (\cref{sec:offset}).

\emph{What does a low score mean?} A cell can receive little attribution because no block identifies evidence there, or because different blocks assign opposing contributions that cancel. A signed sum cannot distinguish these cases. We therefore measure agreement across blocks and suppress cells whose contributions are inconsistent (\cref{sec:composition}).

The resulting components are combined into a single spatiotemporal attribution tensor, which provides both spatial and temporal views (\cref{sec:map}).

\paragraph{Preliminaries.}
A $T$-frame video $x\in\mathbb{R}^{3\times T\times H\times W}$ is embedded into $n$ spatial
tokens per frame plus a class token, a sequence of length $L=1+Tn$. Tokens are ordered
segment-major, with the class token at index $0$. Each block applies a bidirectional selective
SSM. In one directional branch, a depthwise causal convolution first gives
$\mathbf{u}_s=\operatorname{SiLU}(\operatorname{Conv}(\mathbf{x})_s)$. The selective SSM then
updates a state and reads it out:
\begin{equation}
\begin{aligned}
\mathbf{h}_s
&=
\bar{\mathbf{a}}_s \odot \mathbf{h}_{s-1}
+
\left(\Delta_s B_s^{\!\top}\right)\mathbf{u}_s,
\\
y_{s,c}
&=
\sum_n C_{s,n}h_{s,c,n}
+
D_cu_{s,c}.
\end{aligned}
\label{eq:prelim_scan}
\end{equation}
Here $\mathbf{h}_{-1}=\mathbf{0}$, the quantities $\Delta_s$, $B_s$ and $C_s$ depend on the
input, and $\bar a_{s,c,n}=\exp(\Delta_{s,c}A_{c,n})$. The output is then gated and projected
pointwise, which adds no token interactions. The forward branch processes the class token
first, so there it cannot see later tokens. The backward branch processes it last, so there
it aggregates the whole sequence. We call this position $s^\ast=L-1$ the \emph{readout
position}. A token at original position $p$ sits at scan position $s=L-1-p$ in the backward
branch. We anchor the decomposition at the readout state, not at an intermediate state.

\paragraph{Readout-anchored decomposition.}
Unrolling \cref{eq:prelim_scan} to the readout position writes the state as a sum over scan
positions:
\begin{equation}
\mathbf{h}_{s^\ast}
=
\sum_{s=0}^{s^\ast}
\mathbf{q}_s,
\qquad
\mathbf{q}_s
=
\left[
\prod_{r=s+1}^{s^\ast}\bar{\mathbf{a}}_r
\right]
\odot
\left(\Delta_s B_s^{\!\top}\right)\mathbf{u}_s.
\label{eq:state_contribution}
\end{equation}
Each $\mathbf{q}_s$ is the exact contribution of scan position $s$ to the readout state: the
input injected at $s$, decayed by every step between $s$ and the readout.
\Cref{app:state_contribution} writes it out for the diagonal dynamics of VideoMamba, where the
decay is a single exponential of accumulated time steps. Ignoring the direct input term, the
readout is $\mathbf{y}_{s^\ast}=\mathbf{C}_{s^\ast}\mathbf{h}_{s^\ast}
=\sum_{s}\mathbf{C}_{s^\ast}\mathbf{q}_s$, so the contribution of position $s$ to output channel
$c$ is $\varphi_{c,s}=[\mathbf{C}_{s^\ast}\mathbf{q}_s]_c$.

\paragraph{From state contributions to logit contributions.}
To score token $p$ for class $k$, we pass its contribution through the pointwise gate and the
output projection of block $\ell$ and project the result onto a readout direction
$\mathbf{g}^{(\ell)}\in\mathbb{R}^{D}$. This gives a signed score per token and block,
\begin{equation}
\phi_p^{(\ell)}
=
\sum_c
w_c^{(\ell)}
\varphi_{c,L-1-p}^{(\ell)},
\qquad
w_c^{(\ell)}
=
\left\langle
\mathbf{g}^{(\ell)},
W_{\mathrm{out}}^{(\ell)}[:,c]
\right\rangle
\operatorname{SiLU}
\!\left(z_{c,s^\ast}^{(\ell)}\right),
\label{eq:channel_weight}
\end{equation}
and the token score is their sum over the $N_L$ blocks,
$\phi_p=\sum_{\ell=1}^{N_L}\phi_p^{(\ell)}$. The target class enters only through $w_c^{(\ell)}$, and everything
downstream is linear in $\mathbf{g}^{(\ell)}$, so the choice of that direction decides whether the
decomposition is faithful to the classifier. The decomposition within each block is exact, but
$\phi_p$ is a first-order estimate of the token's contribution to the final logit: later blocks
transform the output of earlier ones, and \our{} treats the residual path locally as an identity
instead of computing the Jacobian of the remaining network. We call $\phi_p$ the \emph{logit
estimator}.

\subsection{The Readout Direction Through the Final Normalisation}
\label{sec:readout_direction}

The obvious direction is the classifier row, $\mathbf{g}^{(\ell)}=W_{\mathrm{head}}[k]$. With it,
$\sum_\ell\sum_p\phi_p^{(\ell)}$ reproduces the logit exactly, so it decomposes the logit value. It is not the direction the logit is sensitive to. The classifier reads the
residual stream $\mathbf{r}$ through a normalisation without bias, so the logit depends on the
direction of $\mathbf{r}$ and not on its length, and its gradient is the component of
$\mathbf{w}=W_{\mathrm{head}}[k]\odot\boldsymbol{\gamma}_f$ orthogonal to the stream. The
component thus removed is not small: over the logit range we observe, it is $32$--$44\%$ of the
readout direction. A token that only rescales the residual stream leaves the logit unchanged and
should get no credit. With the classifier row it gets credit.

One gradient is still not enough, because the contributions we decompose are not infinitesimal.
At the median, one block displaces the residual stream by $23.8\%$ of its length, and the first
block by $97\%$. A Jacobian evaluated once, at the final state, is therefore worst at the early
blocks, where the decomposition needs it most. The stream accumulates as
$\mathbf{r}^{(\ell)}=\mathbf{r}^{(\ell-1)}+\mathbf{m}^{(\ell)}$, so a block's contribution to the
normalised logit $g$ is an exact path integral over its own increment. We use the path-averaged
gradient as the readout direction of block $\ell$:
\begin{equation}
\mathbf{g}^{(\ell)}
\;=\;
\int_0^1
\nabla g\!\left(\mathbf{r}^{(\ell-1)}+s\,\mathbf{m}^{(\ell)}\right)\mathrm{d}s .
\label{eq:path_direction}
\end{equation}
This adds no model evaluation. $\mathbf{r}^{(\ell)}$ and $\mathbf{m}^{(\ell)}$ are
$D$-dimensional vectors read from the same forward pass, and $\mathbf{g}^{(\ell)}$ depends only on
block $\ell$, so each block's state can still be released before the next block is processed.
\Cref{app:path_direction} gives the closed form of $\nabla g$, the derivation, the quadrature and
three numerical checks: exact conservation, the endpoint projection as the limiting case, and the
angle between the two directions as a function of depth.

\subsection{Correcting the Convolution-Induced Offset}
\label{sec:offset}

The decomposition is exact for the signal entering the SSM, but a depthwise causal convolution
produces that signal. The signal at a scan position therefore mixes in neighbouring tokens,
which shifts the map systematically. With segment-major ordering, this happens mainly within a
frame. We hand the raw scores back along the convolution taps:
\begin{equation}
\Phi^\star_p
=
\sum_j \omega_j \Phi_{p-j},
\qquad
\sum_j\omega_j=1.
\label{eq:offset_correction}
\end{equation}
The weights $\omega$ come from the convolution itself: the kernel of the backward branch, per
channel and normalised to unit mass. We apply \cref{eq:offset_correction} to each block's
credit before summing over channels, not to the finished map. The taps stay within a
frame, because crossing a frame boundary would move credit between unrelated content. The
kernel differs between blocks. Using all of it strictly generalises the one-patch shift this
correction started as. \Cref{app:offset_detail} gives the measured kernels, the resulting
sub-patch offset and three controls.

\subsection{Channel Truncation and Cancellation Weighting}
\label{sec:composition}

Two post-processing operations modify the map without changing the underlying decomposition. We
ablated each on the same base before combining them (\cref{app:ablation_grid}).

\paragraph{Channel truncation.}
The weights $w_c^{(\ell)}$ of \cref{eq:channel_weight} are concentrated: most of their mass sits
in a few of the $d=2D$ inner channels. Per block, we keep the $\kappa=96$ channels of largest
$\lvert w_c^{(\ell)}\rvert$ and zero the rest. The kept set depends on the direction: switching
from the classifier row to the path-averaged direction of \cref{eq:path_direction} changes $19\%$
of it. The two choices interact and must be tuned together.

\paragraph{Cancellation weighting.}
A cell can score low because no block gives it evidence, or because blocks give it evidence that
cancels. The signed sum cannot tell these apart. We measure agreement as $\mathrm{coh}_p\in[0,1]$,
the magnitude of the signed sum over the sum of magnitudes, and down-weight the cell by
$\mathrm{coh}_p^{\beta}$ with $\beta=2$ (\cref{app:composition_detail}). A cell whose blocks agree
is unchanged, and a cell whose blocks cancel is suppressed. Setting $\beta=0$ recovers the
uncorrected map bitwise, which we use as an identity check. Together with the truncation above,
this is what makes the maps sparse.

\subsection{Spatiotemporal Readout}
\label{sec:map}

After block aggregation, offset correction and the operations of \cref{sec:composition}, we
reshape the scores onto the patch grid, $\Phi\in\mathbb{R}^{T\times H_p\times W_p}$. $\Phi_t$ is
the spatial map of frame $t$. The frame-level importance is its normalised spatial mean:
\begin{equation}
\tau_t
=
\frac{1}{n_{hw}}
\sum_{i,j}\Phi_{t,i,j}
\Bigm/
\left(\frac{1}{n_{hw}}\sum_{i,j}\pi_{t,i,j}\right)^{\gamma},
\qquad
\gamma=\tfrac{1}{2},
\label{eq:temporal}
\end{equation}
where $\pi$ is accumulated alongside $\Phi$ in the same pass: it is
\cref{eq:state_contribution} evaluated with the token content $\mathbf{u}$ and all signs removed,
per block and summed, so it measures how much \emph{could} reach the readout from a position,
irrespective of what is there and of the target class. Dividing by it removes the part of a
frame's score that any target would receive; $\gamma=0$ leaves the plain spatial average. Both
views are read off the same $\Phi$, the temporal one after this normalisation, so the spatial map
is not re-derived from a second estimator.



\section{Experiments and Results}
\label{sec:experiments}

We evaluate \our{} from five complementary perspectives: computational efficiency, faithfulness, localisation, attribution concentration and qualitative behaviour. We first quantify the computational cost required to obtain an attribution map, and then compare the ability of different attribution methods to identify evidence used by the model. We further evaluate whether \our{} identify the temporal extent of the action being performed and compactness in both temporal and spatial meaning of \our{}'s explanations. Finally, we provide qualitative comparisons with attribution-based baselines and content-free controls.
\begin{wraptable}{r}{0.7\linewidth}
\caption{\textbf{Computational cost of attribution.} Measurements for the \textsc{VideoMamba-M} backbone on a single RTX~4090 with batch size~1, averaged over 25 runs after warm-up. Time and memory report the additional cost of computing an attribution map beyond a standard classification forward pass, which takes 37,ms and 0.51,GB. Fwd./Bwd. denotes the number of forward and backward passes required by each method. The $^{*}$ variants use optimized implementations: fused selective-scan kernels for MambaLRP and streaming accumulation for \our{}.}

\label{tab:efficiency}
\centering\small
\setlength{\tabcolsep}{5pt}
\begin{tabular}{@{}lccrrr}
\toprule
Method & Fwd.\ & Bwd.\ & Time (ms) $\downarrow$ & Mem.\ (GB) $\downarrow$ & Aux.\ mem. \\
\midrule
Layer-CAM                 & 1  & 1  & $194$      & $5.67$ & $\mathcal{O}(N_L L)$ \\
IntGrad                   & 16 & 16 & $2\,017$   & $5.47$ & $\mathcal{O}(N_L L)$ \\
MambaLRP                  & 1  & 1  & $19\,807$  & $6.08$ & $\mathcal{O}(N_L L)$ \\
MambaLRP$^{*}$            & 1  & 1  & $143$      & $7.01$ & $\mathcal{O}(N_L L)$ \\
\midrule
\rowcolor{red!10}
\our{} (ours)             & \textbf{1} & \textbf{0} & $\mathbf{125}$ & $3.47$ & $\mathcal{O}(L)$ \\
\rowcolor{red!10}
\our{} (ours)$^{*}$       & \textbf{1} & \textbf{0} & $172$ & $\mathbf{0.06}$ & $\mathcal{O}(L)$ \\
\bottomrule
\end{tabular}
\end{wraptable}
\vspace{-\baselineskip}
\paragraph{Experimental Setup.}
We evaluate \our{} across the full family of pretrained \textsc{VideoMamba} models on three video benchmarks: Something-Something V2 (SSv2), Kinetics-400, and COIN. We compare against Layer-CAM, Integrated Gradients (IntGrad), and MambaLRP, together with two content-independent controls, Centre and Random. All methods are evaluated under the same attribution target and model predictions. We assess faithfulness at temporal, spatial, and spatio-temporal levels using the insertion--MoRF AUC gap, $\mathrm{AUC}_{\mathrm{Ins}}-\mathrm{AUC}_{\mathrm{MoRF}}$. Unless stated otherwise, masking replaces selected pixels with the dataset mean. Localisation is evaluated using temporal annotations from COIN, which specify the temporal extent of individual procedure steps. We report the fraction of clips whose highest-weighted frame falls within the annotated step, the fraction of positive attribution mass contained within the step, and temporal intersection-over-union (tIoU). We additionally measure computational cost in terms of wall-clock time and peak GPU memory. Qualitative comparisons are reported alongside the quantitative evaluation to illustrate differences in the spatial and temporal structure of the resulting attribution maps.

\paragraph{Computational Efficiency}
The main practical advantage of \our{} is that obtaining an attribution map requires a single model forward and no backward pass. For a fixed model size and number of integration steps (for IntGrad),
all compared methods have time complexity linear in sequence length, but differ in the number of forward and backward evaluations.
Table~\ref{tab:efficiency} reports their computational requirements. On \textsc{VideoMamba-M}, \our{} produces a map in $88$\,ms, against
$125$\,ms for Layer-CAM and $0.99$\,s for IntGrad. The released MambaLRP code takes $14.9$\,s
per map. That is a property of its implementation, not of the propagation rule, so we
re-implemented the rule on the fused selective-scan kernel. 

The memory usage difference follows from the computation graph. The per-block terms of
\cref{eq:channel_weight} are independent, and the direction of \cref{eq:path_direction} depends
only on its own block, so each block's state can be released before the next is processed. The chain rule couples the blocks, so relevance propagation must keep all their activations. Accumulating block by block (\our{}$^{*}$) costs $172$\,ms at VideoMamba-Medium and lowers the peak from $3.478$ to $0.06$\,GB, with a bit-identical map. The fused MambaLRP baseline peaks at $7.01$\,GB. Full results for all \textsc{VideoMamba} backbones are reported in Table~\ref{tab:efficiency_full}.

\begin{wraptable}{r}{0.7\linewidth}
\vspace{-1.1\baselineskip}
\caption{\textbf{Temporal localisation on COIN.} Results on 300 test clips for each VideoMamba backbone. Ti, S, and M denote the Tiny, Small, and Medium backbones, respectively. Peak in step is the fraction of clips for which the most-attributed frame falls within the annotated procedure step; Mass in step is the fraction of positive attribution mass within the step; and tIoU is the temporal intersection-over-union between the top-$25\%$ attributed frames and the annotated step. Chance denotes the expected value for a uniformly distributed temporal attribution. Bold indicates the best attribution method in each column.}

\label{tab:temporal_localisation}
\centering\small
\setlength{\tabcolsep}{2.8pt}
\begin{tabular}{@{}lccccccccc@{}}
\toprule
Method & \multicolumn{3}{c}{peak in step $\uparrow$} & \multicolumn{3}{c}{mass in step $\uparrow$} & \multicolumn{3}{c}{tIoU $\uparrow$} \\
\cmidrule(lr){2-4}\cmidrule(lr){5-7}\cmidrule(lr){8-10}
 & \textsc{Ti} & \textsc{S} & \textsc{M} & \textsc{Ti} & \textsc{S} & \textsc{M} & \textsc{Ti} & \textsc{S} & \textsc{M} \\
\midrule
\textcolor{gray}{Centre} & \textcolor{gray}{0.073} & \textcolor{gray}{0.073} & \textcolor{gray}{0.073} & \textcolor{gray}{---} & \textcolor{gray}{---} & \textcolor{gray}{---} & \textcolor{gray}{0.106} & \textcolor{gray}{0.106} & \textcolor{gray}{0.106} \\
\textcolor{gray}{Random} & \textcolor{gray}{0.503} & \textcolor{gray}{0.443} & \textcolor{gray}{0.520} & \textcolor{gray}{0.501} & \textcolor{gray}{0.500} & \textcolor{gray}{0.500} & \textcolor{gray}{0.181} & \textcolor{gray}{0.173} & \textcolor{gray}{0.173} \\
Layer-CAM & 0.370 & 0.177 & 0.203 & 0.454 & 0.344 & 0.377 & 0.173 & 0.160 & 0.162 \\
IntGrad & 0.470 & 0.460 & 0.487 & 0.500 & 0.502 & 0.500 & 0.175 & 0.178 & 0.181 \\
MambaLRP & 0.503 & 0.473 & 0.470 & 0.565 & 0.550 & 0.557 & 0.223 & 0.219 & 0.215 \\
\rowcolor{red!10}
\our{} (ours) & \textbf{0.593} & \textbf{0.533} & \textbf{0.583} & \textbf{0.606} & \textbf{0.595} & \textbf{0.597} & \textbf{0.236} & \textbf{0.238} & \textbf{0.238} \\
\bottomrule
\end{tabular}
\end{wraptable}
\vspace{-0.32cm}\paragraph{Faithfulness.}
We evaluate whether an attribution method ranks the evidence used by the model above alternative regions of the input. We measure faithfulness using the insertion--MoRF AUC gap, computed independently at temporal, spatial, and spatio-temporal levels. A larger gap indicates that the attributed regions are more informative for the target prediction. Inserting highly attributed regions increases the prediction more rapidly, while removing them causes a larger decrease. Table~\ref{tab:faithfulness} shows that \our{} achieves the strongest temporal faithfulness on both COIN and Kinetics-400, while remaining competitive with MambaLRP at the spatial and joint levels. This result indicates that the proposed decomposition is particularly effective at identifying temporally relevant evidence, without substantially sacrificing spatial or spatio-temporal faithfulness. The relative performance across attribution methods is consistent across the evaluated backbone sizes, despite differences in the underlying model capacity. The complete faithfulness results, including insertion and deletion curves and the alternative ROAD evaluation, are provided in Appendix~\ref{app:additional_faithfulness}.

  \begingroup
  \renewcommand{\caption}[2][]{\READorigcaption{\textbf{Faithfulness} at VideoMamba-S: the insertion$-$MoRF AUC gap, $\mathrm{AUC}_{\mathrm{Ins}}-\mathrm{AUC}_{\mathrm{MoRF}}$, at the
temporal, spatial and spatio-temporal levels (higher is
better). Bold marks the best result per column. The grey rows are controls that see neither the model nor the target class.}}%
  \begin{table}[t]
\centering
\caption{
Faithfulness comparison across Something-Something v2 (SSv2), Kinetics-400 (K-400) and COIN.
We report the insertion--MoRF AUC gap at temporal, spatial, and
spatio-temporal levels; higher is better. The score is computed as
$\mathrm{AUC}_{\mathrm{ins}}-\mathrm{AUC}_{\mathrm{MoRF}}$ using the
same evaluation protocol for all methods. Bold marks the best value in each column among the attribution methods; the two grey rows are controls that see neither the model nor the target class, and they bound what a map can score without using either.}.

\label{tab:faithfulness}
\setlength{\tabcolsep}{3.8pt}
\begin{tabular}{@{}lccc ccc ccc@{}}
\toprule
& \multicolumn{3}{c}{Temporal $\uparrow$}
& \multicolumn{3}{c}{Spatial $\uparrow$}
& \multicolumn{3}{c}{Spatio-temporal $\uparrow$} \\
\cmidrule(lr){2-4}
\cmidrule(lr){5-7}
\cmidrule(lr){8-10}
Method
& \textit{SSv2} & \textit{K-400} & \textit{COIN}
& \textit{SSv2} & \textit{K-400} & \textit{COIN}
& \textit{SSv2} & \textit{K-400} & \textit{COIN} \\
\midrule
\textcolor{gray}{Centre}
& \textcolor{gray}{$-0.010$} & \textcolor{gray}{$-0.010$} & \textcolor{gray}{$-0.003$}
& \textcolor{gray}{$0.101$} & \textcolor{gray}{$0.053$} & \textcolor{gray}{$0.009$}
& \textcolor{gray}{$0.100$} & \textcolor{gray}{$0.053$} & \textcolor{gray}{$0.006$} \\

\textcolor{gray}{Random}
& \textcolor{gray}{$-0.025$} & \textcolor{gray}{$-0.024$} & \textcolor{gray}{$-0.018$}
& \textcolor{gray}{$-0.031$} & \textcolor{gray}{$-0.029$} & \textcolor{gray}{$-0.032$}
& \textcolor{gray}{$-0.031$} & \textcolor{gray}{$-0.029$} & \textcolor{gray}{$-0.032$} \\

Layer-CAM
& $0.095$ & $0.049$ & $0.063$
& $0.281$ & $0.241$ & $0.209$
& $0.302$ & $0.258$ & $0.243$ \\

IntGrad
& $0.081$ & $0.074$ & $0.138$
& $0.205$ & $0.236$ & $0.250$
& $0.217$ & $0.245$ & $0.265$ \\

MambaLRP
& $\mathbf{0.184}$ & $0.139$ & $0.253$
& $\mathbf{0.451}$ & $\mathbf{0.536}$ & $\mathbf{0.564}$
& $\mathbf{0.474}$ & $\mathbf{0.547}$ & $\mathbf{0.610}$ \\

\rowcolor{red!10}
\our{} (ours)
& $0.154$ & $\mathbf{0.144}$ & $\mathbf{0.298}$
& $0.364$ & $0.433$ & $0.467$
& $0.390$ & $0.452$ & $0.547$ \\

\bottomrule
\end{tabular}
\end{table}
  \endgroup

\vspace{-0.3cm}\paragraph{Temporal Localisation.}
We evaluate whether attribution methods identify the temporal extent of the action being performed. COIN provides temporal annotations for each procedure step, allowing us to compare the attributed temporal profile with the corresponding annotated interval. We report three measures: the fraction of clips whose highest-weighted frame falls inside the annotated step (Peak in step), the fraction of positive attribution mass contained within the step (Mass in step), and the temporal IoU between the top-$25\%$ frames and the annotated interval (tIoU). Results are shown in \cref{tab:temporal_localisation}. \our{} consistently achieves the highest values across all three metrics and all evaluated backbone sizes. In particular, its peak falls within the annotated step in $53$--$59\%$ of clips, while $59$--$61\%$ of its attribution mass lies within the step. The Random control remains close to the expected chance level, providing a reference for the temporal localisation task.

\begin{figure}[t]
    \centering
    \includegraphics[width=\linewidth]{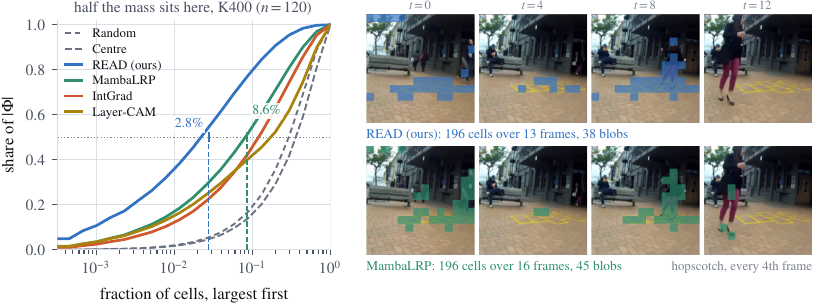}
    \caption{\textbf{Attribution concentration and spatial-temporal coherence.} Left: cumulative attribution mass as a function of the fraction of spatio-temporal cells, averaged over 120 Kinetics-400 clips. Curves further to the left indicate that the attribution mass is concentrated on fewer cells. The dashed horizontal line marks $50\%$ attribution mass. Right: attribution maps for the same clip at a $\frac{1}{16}$ attribution budget, shown every second frame. \our{} concentrates attribution on fewer frames and more spatially compact regions than the compared attribution method, while both methods select the same number of cells. Full quantitative results across SSv2, Kinetics-400, and COIN are reported in \cref{app:concentration}.}
    \label{fig:skupienie}
    \vspace{-0.4cm}
\end{figure}

\vspace{-0.3cm}\paragraph{Attribution Concentration.}
Beyond faithfulness and localisation, we examine how spatially and temporally concentrated the resulting attribution maps are. We measure concentration by the fraction of spatio-temporal cells required to accumulate half of the total absolute attribution mass (Mass@$\frac{1}{2}$), and coherence by the fraction of selected cells that are adjacent to another selected cell (Contiguity). Across SSv2, Kinetics-400, and COIN, \our{} consistently produces more concentrated and coherent maps than the attribution baselines. At the $\frac{1}{16}$ attribution budget, this concentration translates into explanations that focus on fewer frames and spatially compact regions, as illustrated in \cref{fig:skupienie}. We
report the complete quantitative results in \cref{app:concentration}.
\begin{figure}[ht]
    \centering
    \includegraphics[width=\linewidth]{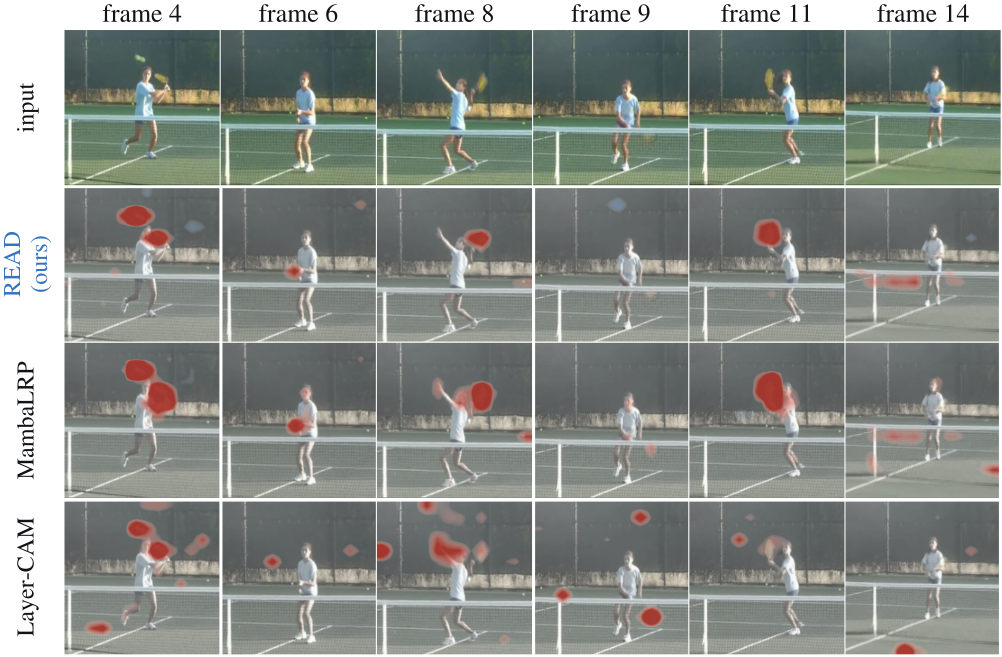}
    \caption{\textbf{Qualitative comparison on Kinetics-400}, class ``playing tennis'',
    correctly classified by VideoMamba-S. The nine of the $16$ frames (index above each
    column); rows show the input, \our{}, three baselines. One
    diverging colormap for all maps (red: evidence for the predicted class; blue: against).}
    \label{fig:qualitative_comparison}
\end{figure}
report the complete quantitative results in \cref{app:concentration}.

\vspace{-0.3cm}\paragraph{Qualitative Comparison.}
\Cref{fig:qualitative_comparison} places \our{} beside every baseline on a \emph{playing tennis}
clip that \textsc{VideoMamba-S} classifies correctly; \cref{app:qualitative} gives more examples,
including two further Kinetics-400 clips (\cref{fig:violins}). In the swing frames (4, 8, 11),
\our{} places its mass on the racket and the striking arm and leaves the empty court largely
unattributed. All frames are read out of one decomposition, so the maps change with the action
instead of repeating a fixed pattern. Layer-CAM responds to scattered background fragments with little temporal structure.

The temporal attribution provides an additional view of how the methods distribute relevance across the video. In particular, \our{} tends to assign temporal importance to frames that correspond to meaningful changes in the visual content, rather than consistently favouring a fixed position in the sequence. This behaviour is further illustrated by additional qualitative examples and temporal attribution visualisations in the appendix. There, we provide temporal importance profiles alongside the corresponding spatial attribution maps, allowing the temporal and spatial patterns of the explanations to be examined jointly. We additionally analyse the tendency to assign the highest temporal relevance to the first frame. Across $300$ Kinetics-400 validation clips, frame~$0$ carries the maximum of $\tau_t$ for $86.7\%$ of clips under Layer-CAM and $67.0\%$ under MambaLRP, compared with $28.0\%$ for \our{}. Across all nine dataset--backbone combinations (\cref{tab:first_frame}), the attribution baselines exhibit a systematic first-frame preference, whereas \our{} is indistinguishable from chance on all SSv2 and COIN combinations. On Kinetics-400, \our{} retains a smaller residual bias ($12.3$--$28.0\%$ compared with a chance level of $6.2\%$), which is also the dataset where its temporal faithfulness advantage is largest (\cref{tab:faithfulness}). Additional temporal attribution examples and the corresponding temporal importance profiles are provided in \cref{app:qualitative}.

\section{Conclusion}
\label{sec:conclusion}

We introduced \our{}, a parameter-free attribution method for selective state-space video models that reads the evidence for a prediction directly from the recurrence. Once the input-dependent coefficients of the scan are fixed, the state at the classifier's readout decomposes exactly into per-token contributions. \our{} projects these contributions onto a path-averaged class direction that accounts for the final normalisation, returns each block's credit along the taps of the causal convolution, and suppresses cells on which blocks cancel. The result is a single spatiotemporal map, from which both the spatial and the temporal view are read. It is obtained in one forward pass, without backpropagation, and with auxiliary memory
independent of model depth. Across three datasets and three backbone sizes, \our{} achieves the strongest temporal faithfulness on Kinetics-400 and COIN, the best temporal localisation on COIN at every backbone size, and more concentrated and contiguous maps than the attribution baselines,
while remaining competitive with MambaLRP at the spatial and spatio-temporal levels. It also largely avoids the first-frame bias that every evaluated baseline inherits from the scan geometry. More broadly, our results indicate that in selective SSMs the recurrence is not only an efficient mechanism for inference but also a direct and inexpensive source of evidence, and that
faithful attribution can follow the structure of the computation instead of approximating it from outside.

\paragraph{Limitations.} The relevance decomposition is exact within each block, but the contribution to the final logit is approximated across blocks using a first-order propagation rule. The resulting attribution is therefore not an exact decomposition of the final logit across the full network. A second limitation arises from the directional scan in the underlying Mamba architecture, which can result in elevated attribution to early frames. Our formulation substantially reduces this effect on SSv2 and COIN, but does not fully eliminate it on Kinetics-400. Finally, our evaluation is limited to VideoMamba classifiers with a class-token readout and does not establish whether the same construction applies to other readout mechanisms or selective state-space architectures.

\subsection*{AI use statement}
We used generative AI tools as assistive tools during the preparation and implementation of this work. In writing, we used large language models to improve the clarity and fluency of author-written text, refine phrasing, shorten or restructure passages, and improve the overall presentation of the paper. In the implementation of the experiments, generative AI tools were also used as coding assistants to help write, refactor, and debug code, including experiment scripts and utilities for processing measurements and generating tables and figures. All generated code was reviewed, tested, and adapted by the authors.

Generative AI was not used to invent the proposed method, formulate its underlying ideas, define the method or its components, establish the mathematical formulation, choose the research questions, design the experimental methodology, define the evaluation metrics, or interpret the experimental results. The scientific decisions and conclusions presented in this paper were made by the authors. The authors reviewed and verified the final text, code, experimental results, and figures, and take full responsibility for the content of this work.

\subsection*{Reproducibility statement}

We provide the details necessary to reproduce our method and experiments throughout the paper and supplementary material. The method is described in \cref{sec:method}, with additional methodological details provided in \cref{app:method_details}. The experimental setup, datasets, training and evaluation procedures, and implementation details are described in \cref{app:experimental_details}. We additionally share the source code, experiment configurations, and scripts used to reproduce the reported results and generate the tables and figures in the supplementary material. All experiments use publicly available datasets and pretrained \textsc{VideoMamba} models.

\bibliography{iclr2027_conference}
\bibliographystyle{iclr2027_conference}

\appendix
\crefalias{section}{appendix}
\crefalias{subsection}{appendix}
\section*{Appendix}
\Cref{app:experimental_details} describes the three datasets, including the part of COIN that
can no longer be retrieved and that bounds every COIN number we report, and explains which
architecture-specific baseline we compare against. \Cref{app:method_details} gives the
derivations and numerical checks behind \cref{sec:method}: the closed form of the state
contribution, the readout direction, the measurement behind the convolution correction, the
temporal normalisation and an activation-patching control. \Cref{app:ablations} gives the
ablations. \Cref{app:additional_faithfulness} gives the evaluation protocol, the metric
definitions and the quantitative results that do not fit in the main text. \Cref{app:cost} gives
the full cost measurement and the mechanism behind each starred variant. \Cref{app:qualitative}
gives further qualitative comparisons on all three datasets.

\section{Additional Implementation Details}
\label{app:experimental_details}

\subsection{Datasets}
\label{app:datasets}

We evaluate all three datasets with the released VideoMamba checkpoints, at the frame count each
checkpoint was trained for, under the protocol of \cref{app:additional_faithfulness}.

\paragraph{Something-Something~v2} \citep{goyal2017something} has $174$ classes defined by the
manipulation rather than by the object, so a class usually cannot be read off a single frame. We
use the validation split ($24\,777$ clips) at $16$ frames. Through the Something-Else
extension \citep{materzynska2020something}, it is the only one of our datasets with frame-level
box annotations. We therefore measure spatial localisation here and only here: $300$ clips, of
which $4\,645$ of $4\,800$ frames carry a box.

\paragraph{Kinetics-400} \citep{kay2017kinetics} has $400$ classes. We use the validation split
at $16$ frames, of which $19\,881$ of $19\,906$ clips still resolve. One detail fails silently
and is worth recording. The class order of the released checkpoint is a permutation of the
alphabetical label list. With the alphabetical order, the model runs normally but at chance
accuracy. We load the permutation shipped with the checkpoint and verify it by top-1 accuracy
before computing any attribution.

\paragraph{COIN} \citep{tang2019coin} has $180$ procedural tasks. It is the only one of our
datasets that annotates step boundaries, in seconds. We use the testing split at $32$ frames.
\textbf{The full testing split is not available.} COIN is distributed as a list of YouTube
identifiers, not as video files, and much of that list no longer resolves. Of the $10\,526$
identifiers in our download list, $7\,587$ were unavailable. This left $2\,939$ videos,
$2\,233$ in the training split and $\mathbf{706}$ in the testing split. Every COIN number in this
paper is computed on those $706$ clips and is not directly comparable with a number computed on
the complete official split. Within them, the faithfulness measurements use $689$--$691$ clips at
all three backbone sizes (of $706$ requested; the rest fail to decode), the frame budgets of
\cref{fig:budzet} use $491$--$500$, and the temporal localisation of
\cref{tab:temporal_localisation} uses $300$. All $706$ clips carry step annotations, so
availability is the only restriction on the temporal measurement.

\subsection{On the Choice of the Architecture-Specific Baseline}
\label{app:baseline_choice}
The natural second architecture-specific baseline is the implicit-attention formulation of
\citet{ali2025hidden}. We did not use it, for a reason that lies in the sequence length rather
than in the method. The formulation makes the scan explicit as an $L\times L$ operator per inner
channel and scan direction. Its authors note that a selective SSM yields three orders of
magnitude more such matrices than a Transformer. At video sequence lengths this cannot be
materialised. With $L=3137$ for $16$ frames, $d=768$ inner channels and two scan directions, a
single block needs $56$\,GB and the full backbone $1.3$\,TB. At $32$ frames ($L=6273$), the same
figures are $225$\,GB per block and $5.3$\,TB. These are exact counts in \texttt{fp32}, not
estimates. The formulation becomes tractable if one keeps only the row of the implicit attention
at the readout position. That row, however, is exactly what our decomposition computes in
closed form, so the comparison would not be independent. We therefore compare against MambaLRP.
It is defined for the same architecture, released by its authors, and we run it without
modification.

\section{Derivations and Details of \our{}}
\label{app:method_details}

This appendix gives reproduction details that are not needed to follow the main method.

\subsection{The State Contribution in Closed Form}
\label{app:state_contribution}

For the diagonal Mamba dynamics, the channel contribution of \cref{sec:method} is
\begin{equation}
\varphi_{c,s}
=
\sum_{n=1}^{N}
C_{s^\ast,n}
\exp\!\left(
A_{c,n}
\sum_{r=s+1}^{s^\ast}\Delta_{r,c}
\right)
\Delta_{s,c}B_{s,n}u_{s,c}.
\end{equation}
Each term combines the information injected at position $s$, its propagation to the readout and
the readout coefficient $C_{s^\ast,n}$. For original token position $p$, the backward-scan
contribution is $\varphi_{c,L-1-p}$.

\subsection{The Readout Direction: Closed Form, and Why the Endpoint Gradient Is Not Enough}
\label{app:path_direction}

The classifier reads the residual stream through a normalisation without bias:
\begin{equation}
\mathrm{logit}_k
=
\left\langle
W_{\mathrm{head}}[k],\;
\frac{\mathbf{r}}{\mathrm{rms}(\mathbf{r})}\odot\boldsymbol{\gamma}_f
\right\rangle
+ b_k
=
\frac{\langle \mathbf{w},\mathbf{r}\rangle}{\mathrm{rms}(\mathbf{r})} + b_k,
\qquad
\mathbf{w}=W_{\mathrm{head}}[k]\odot\boldsymbol{\gamma}_f,
\label{eq:readout_norm}
\end{equation}
where $\mathbf{r}$ is the residual stream at the readout position and
$\mathrm{rms}(\mathbf{r})=\lVert\mathbf{r}\rVert/\sqrt{D}$. With
$g(\mathbf{r})=\sqrt{D}\,\langle\mathbf{w},\mathbf{r}\rangle/\lVert\mathbf{r}\rVert$, the
gradient is the component of $\mathbf{w}$ orthogonal to the stream:
\begin{equation}
\nabla g(\mathbf{r})
=
\frac{1}{\mathrm{rms}(\mathbf{r})}
\left(
\mathbf{w}-\langle\mathbf{w},\hat{\mathbf{r}}\rangle\,\hat{\mathbf{r}}
\right),
\qquad
\hat{\mathbf{r}}=\mathbf{r}/\lVert\mathbf{r}\rVert.
\label{eq:rms_jacobian}
\end{equation}
The removed component has closed-form length $\lvert\mathrm{logit}_k-b_k\rvert/\sqrt{D}$, which
over the logit range we observe is the $32$--$44\%$ quoted in \cref{sec:readout_direction}.

\Cref{eq:rms_jacobian} is a first-order expansion, exact only for infinitesimal
contributions, and the displacements quoted in \cref{sec:readout_direction} show that ours
are not.

The residual stream accumulates as $\mathbf{r}^{(\ell)}=\mathbf{r}^{(\ell-1)}+\mathbf{m}^{(\ell)}$,
where $\mathbf{r}^{(0)}$ is the class-token embedding and $\mathbf{m}^{(\ell)}$ the contribution of
block $\ell$. The contribution of a block to the normalised logit is then exact:
\begin{equation}
g\!\left(\mathbf{r}^{(\ell)}\right)-g\!\left(\mathbf{r}^{(\ell-1)}\right)
=
\int_0^1
\left\langle
\nabla g\!\left(\mathbf{r}^{(\ell-1)}+s\,\mathbf{m}^{(\ell)}\right),
\mathbf{m}^{(\ell)}
\right\rangle
\mathrm{d}s
=
\left\langle
\bar{\mathbf{g}}^{(\ell)},\mathbf{m}^{(\ell)}
\right\rangle.
\label{eq:path_ftc}
\end{equation}
The averaged gradient is the direction of \cref{eq:path_direction}. We evaluate the integral by
$16$-node Gauss--Legendre quadrature, which agrees with $64$ nodes to $2\times10^{-7}$.

\paragraph{What this buys.}
Three properties follow, and we check each numerically on the frozen model. First, the
decomposition is \emph{exactly conservative} with respect to the readout nonlinearity. Summing
\cref{eq:path_ftc} telescopes to $g(\mathbf{r}^{(N_L)})-g(\mathbf{r}^{(0)})$, with a median
relative error of $7.5\times10^{-7}$ over blocks and clips. The endpoint projection has no such
property: its sum recovers only $24$--$35\%$ of the plain sum. Second, the endpoint projection is
the limiting case. As $\lVert\mathbf{m}^{(\ell)}\rVert\to0$, the path average collapses to
$\nabla g(\mathbf{r}^{(N_L)})$. The angle between the two grows monotonically towards the input,
from $\cos=0.9997$ at the last block to $\cos=0.938$ at the first. Third, the direction is per
block and costs nothing beyond the forward pass. This is what lets the streaming variant of
\cref{app:cost} release each block's state before reading the next.

\subsection{The Convolution Kernel and the Controls on the Offset Correction}
\label{app:offset_detail}

The kernel differs between blocks, and this matters for how the correction should be read. In
the first block of \textsc{VideoMamba-S} it is $[0.09,0.10,0.18,0.63]$, concentrated on the most
recent position, so a single-patch shift is a reasonable approximation there. Averaged over the
$24$ blocks it is $[0.11,0.24,0.34,0.32]$, with half of its mass away from the largest tap. The
centre of mass of the redistribution is $1.16$ patches, which a one-patch shift rounds to $1$.
Using the whole kernel is therefore a strict generalisation of that shift, not an alternative.

The ablations support the generalisation. Paired over $2000$ clips, the per-channel kernel
improves on the one-patch shift by $+7.6\sigma$ on SSv2, $+16.3\sigma$ on Kinetics-400 and
$+6.3\sigma$ on COIN. Three controls bound the explanation. Widening the kernel with a Gaussian
degrades the result monotonically ($-5.4\sigma$ to $-10.4\sigma$), so the gain is not smoothing.
A uniform kernel of the same width hurts on Kinetics-400, so the gain is not the four-tap width
alone. Keeping the sign of the kernel instead of its magnitude is catastrophic: absolute
faithfulness drops to $-0.087$, below a random map, with sign agreement at chance. The correction
is thus about \emph{where} credit belongs, not about reproducing the convolution as a signed
operator. One control does not separate. Permuting the kernels between channels matches the
true assignment on SSv2. The correction exploits the shape of the redistribution, not the
pairing of a kernel with its channel.

\subsection{Cancellation Weighting}
\label{app:composition_detail}

Agreement between blocks is the magnitude of the signed sum over the sum of magnitudes, and the
cell is down-weighted by a power of it:
\begin{equation}
\mathrm{coh}_p
=
\frac{\bigl\lvert\sum_{\ell,c} w_c^{(\ell)}\varphi_{c,L-1-p}^{(\ell)}\bigr\rvert}
     {\sum_{\ell,c} \bigl\lvert w_c^{(\ell)}\varphi_{c,L-1-p}^{(\ell)}\bigr\rvert}
\in[0,1],
\qquad
\Phi_p \leftarrow \Phi_p\,\mathrm{coh}_p^{\beta},
\label{eq:coherence}
\end{equation}
with $\beta=2$. The ratio is taken over blocks and channels jointly, so a cell is suppressed both
when two blocks disagree about it and when the channels within one block do. Setting $\beta=0$
leaves $\Phi$ bitwise unchanged, which is the identity check we run whenever the weighting is
modified, and $\kappa$ and $\beta$ were tuned together on the grid of \cref{app:ablation_grid}.

\subsection{Why the Temporal View Is Normalised}
\label{app:temporal_norm}

The two views measure different quantities. A frame can carry a large summed score because its tokens
have large magnitude, independently of the class. Dividing by the prior removes the part of the
frame score that any target would receive. We use one value of $\gamma$ for all three datasets.
This was not always possible. Without the readout direction of \cref{sec:readout_direction}, the
best $\gamma$ differed between datasets, which would have made it a threshold fitted per
benchmark rather than part of the method.

\subsection{Which Field Carries the Prediction? An Activation-Patching Control}
\label{app:gate}

The decomposition of \cref{sec:method} attributes through the selective scan and treats the
pointwise gate as a per-channel scaling. This is a modelling choice, so we test what each field
of a Mamba block determines. We transplant fields between clips: the content of clip $B$ is
pushed through fields frozen from clip $A$, and we ask which frozen fields make the logits follow
$A$. ``Gate'' means four different things in this architecture, and we separate them: $\Delta$,
the discretisation step that carries Mamba's selectivity; $B$ and $C$, the input and
state-readout projections; and $z$, the pointwise multiplicative gate inherited from GLU, which
does no mixing along time.

Two controls validate the measurement. Freezing every field from the \emph{same} clip must
reproduce its logits, and it does, to $1.1\times10^{-6}$ on SSv2 and $1.9\times10^{-6}$ on
Kinetics-400. Freezing nothing must follow $B$ exactly, and it gives a correlation of $1.000$
with $B$. A result that moved either control would be an implementation error, not a finding.

\Cref{tab:gate} gives the rest, and the split is sharp. Freezing $z$ alone transfers prediction
identity. The logits correlate at $0.98$ and $0.95$ with the donor clip $A$, and the predicted
class is $A$'s for $88\%$ and $92\%$ of pairs. Freezing the selective dynamics \emph{without}
$z$, that is $\Delta$, $B$ and $C$ together, transfers almost nothing. These are all the fields
specific to a state-space model, yet the correlation with $A$ drops to $0.16$ and $0.08$, and the
predicted class is $A$'s in at most $2\%$ of pairs. The component that decides the prediction is
the one part of the block that is not a scan.

We report this as a control on our modelling choice, not as a claim about Mamba in general. It
uses $40$ clip pairs per dataset and one architecture, and it measures the predicted class rather
than the attribution map. What it supports is narrow but sufficient. The decomposition treats $z$
as a fixed per-channel scaling, and this freezes exactly the field that the intervention shows to
be decisive. The estimator does not silently discard the part of the block that carries the
class.

  \begingroup
  \renewcommand{\caption}[2][]{\READorigcaption{\textbf{Transplanting fields between clips.} The content of clip $B$ is processed
with the listed fields frozen from clip $A$. \emph{corr.\ $A$} and \emph{corr.\ $B$} are the
correlations of the resulting logits with those of each source clip, and \emph{$=A$} is the
percentage of pairs whose predicted class is $A$'s; $40$ pairs per dataset. Freezing $z$ alone
transfers the prediction. Freezing the selective dynamics without it does not. The grey row is
the null control and must follow $B$ exactly, which it does.
}}%
  \begin{table}[tbh]
\caption{\textbf{Transplanting fields between clips.} The content of clip $B$ is processed with
the listed fields frozen from clip $A$. \emph{corr.\ $A$} and \emph{corr.\ $B$} are the
correlations of the resulting logits with those of each source clip, and \emph{$=A$} is the
percentage of pairs whose predicted class is $A$'s; $40$ pairs per dataset. Freezing $z$ alone
transfers the prediction; freezing the selective dynamics without it does not. The grey row is
the null control and must follow $B$ exactly, which it does. Generated by
\texttt{tools/analysis/tabela\_bramki.py}.}
\label{tab:gate}
\centering\small
\setlength{\tabcolsep}{5pt}
\begin{tabular}{@{}l rrr rrr@{}}
\toprule
& \multicolumn{3}{c}{Something-Something~v2} & \multicolumn{3}{c}{Kinetics-400} \\
\cmidrule(lr){2-4}\cmidrule(lr){5-7}
Frozen from $A$ & corr.\ $A$ & corr.\ $B$ & $=A$ (\%)
& corr.\ $A$ & corr.\ $B$ & $=A$ (\%) \\
\midrule
all four & $0.986$ & $0.112$ & $92$ & $0.975$ & $0.055$ & $92$ \\
$z$ only & $0.977$ & $0.123$ & $88$ & $0.946$ & $0.071$ & $92$ \\
$\Delta$, $z$ & $0.985$ & $0.118$ & $92$ & $0.973$ & $0.059$ & $98$ \\
$\Delta$, $B$, $C$ (no $z$) & $0.157$ & $0.261$ & $2$ & $0.075$ & $0.324$ & $0$ \\
$\Delta$ only & $0.143$ & $0.351$ & $2$ & $0.005$ & $0.371$ & $0$ \\
$B$, $C$ & $0.106$ & $0.758$ & $0$ & $0.052$ & $0.773$ & $0$ \\
\textcolor{gray}{\textit{nothing (control)}} & \textcolor{gray}{$0.103$} & \textcolor{gray}{$1.000$} & \textcolor{gray}{$2$} & \textcolor{gray}{$0.045$} & \textcolor{gray}{$1.000$} & \textcolor{gray}{$2$} \\
\bottomrule
\end{tabular}
\end{table}
  \endgroup
\section{Why the temporal axis is where the decomposition helps most.}
The scan geometry makes the first frame a trap. The class token is first in the forward order and
therefore last in the backward order, so a token at position $p$ sits at scan position $L-1-p$:
the tokens of frame~$0$ are the \emph{last} to be injected before the readout and the least
decayed when it is taken. Strip the token content and the signs out of
\cref{eq:state_contribution} and what is left is how much \emph{could} arrive at the readout from
each position --- a purely positional ramp peaking at frame~$0$, a property of the ordering and
not of the video. Any estimator whose score behaves like a magnitude accumulated along the scan
inherits that ramp, and that is what we measure: every baseline places its largest frame relevance
on frame~$0$ above chance in all nine dataset--backbone settings (\cref{tab:first_frame}).

\our{} meets the same geometry, but its score is a \emph{signed} projection onto a class
direction rather than an accumulated magnitude, and the ramp multiplies both signs alike. Pure
Gaussian noise isolates the effect, because an estimator carrying a positional prior must
reproduce the frame ramp on an input that contains nothing. Over $40$ noise clips, the
correlation between frame score and frame index is $+0.07\pm0.04$ for \our{} on
Something-Something~v2, indistinguishable from the Random control at $+0.04\pm0.04$, against
$-0.72\pm0.02$ for MambaLRP. On Kinetics-400 the geometry is not cancelled completely: \our{}
reaches $-0.29\pm0.04$, still well inside MambaLRP's $-0.51\pm0.03$ and Layer-CAM's
$-0.50\pm0.03$. The readout direction buys most of the difference: substituting the endpoint
projection for \cref{eq:path_direction} and changing nothing else moves the prior to
$-0.30\pm0.04$ and $-0.40\pm0.03$. That one dataset retains a prior and the other does not is the
same picture the first-frame test reports independently, where \our{} is indistinguishable from
chance in six of nine settings rather than in all nine.
\section{Ablations}
\label{app:ablations}

We examine the effects of the readout direction and post-processing
choices in \our{}. Each variant is evaluated on the same clips
as the reference configuration, enabling paired comparisons.
The values beneath the reported scores indicate standardised paired
differences relative to \our{}; positive values favour the variant.

\subsection{The Readout Direction}
\label{app:ablation_direction}

\Cref{tab:ablation_direction} compares three readout directions:
the classifier weight vector, the endpoint projection
of \cref{eq:rms_jacobian}, and the path-integrated direction
of \cref{eq:path_direction}.
We evaluate the insertion--MoRF gap under the \textsc{mask} protocol
on 2000 clips per dataset.

The lower block compares the three directions within the full estimator,
keeping convolution correction, channel truncation, cancellation
weighting, and temporal normalisation fixed.
The upper block evaluates the endpoint and path-integrated directions
without post-processing, allowing us to examine its contribution
separately.

Using the classifier weight vector reduces spatial and joint
faithfulness on both datasets compared with the path-integrated
direction. This supports accounting for the effect of final
normalisation when projecting contributions onto the target class.
Path integration also improves on the endpoint projection in five
of the six dataset--level combinations, with similar temporal
performance on Kinetics-400.

The effect of post-processing varies across datasets.
On Something-Something~v2, the path-integrated map without
post-processing achieves higher temporal faithfulness than
the full estimator, with scores of 0.174 and 0.154, respectively.
On Kinetics-400, post-processing improves faithfulness at all
three evaluation levels.
We use the same configuration across datasets to maintain
a consistent attribution procedure.

  \begingroup
  \renewcommand{\caption}[2][]{\READorigcaption{
    \textbf{Ablation of the readout direction.}
    Insertion--MoRF gap under the \textsc{mask} protocol;
    higher is better.
    The upper block omits post-processing.
    The lower block uses the full estimator and varies only
    the readout direction.
}}%
  \begin{table}[tbh]
\caption{\textbf{Ablation of the readout direction.} Insertion$-$MoRF gap, mask protocol,
$2000$ clips per dataset, one run per clip so all comparisons are paired. The upper block is
the direction alone; the lower block is the full estimator --- convolution correction, channel
truncation, cancellation weighting and temporal normalisation --- where the three rows differ
only in the direction. The number beneath each value is the paired sigma against \our{}: positive
means the variant beats what we ship. Generated by
\texttt{tools/analysis/tabela\_ablacja.py}.}
\label{tab:ablation_direction}
\centering\small
\setlength{\tabcolsep}{3pt}
\begin{tabular}{@{}l rrr rrr@{}}
\toprule
& \multicolumn{3}{c}{Something-Something~v2} & \multicolumn{3}{c}{Kinetics-400} \\
\cmidrule(lr){2-4}\cmidrule(lr){5-7}
Direction & temporal & spatial & joint & temporal & spatial & joint \\
\midrule
\multicolumn{7}{@{}l}{\emph{direction only, no post-processing}}\\[1pt]
endpoint projection & \begin{tabular}[t]{@{}r@{}}$0.170$\\[-2.5pt]{\scriptsize(+6.4)}\end{tabular} & \begin{tabular}[t]{@{}r@{}}$0.356$\\[-2.5pt]{\scriptsize(-5.0)}\end{tabular} & \begin{tabular}[t]{@{}r@{}}$0.383$\\[-2.5pt]{\scriptsize(-4.2)}\end{tabular} & \begin{tabular}[t]{@{}r@{}}$0.127$\\[-2.5pt]{\scriptsize(-9.6)}\end{tabular} & \begin{tabular}[t]{@{}r@{}}$0.399$\\[-2.5pt]{\scriptsize(-16.7)}\end{tabular} & \begin{tabular}[t]{@{}r@{}}$0.419$\\[-2.5pt]{\scriptsize(-17.0)}\end{tabular} \\
path-integrated & \begin{tabular}[t]{@{}r@{}}$0.174$\\[-2.5pt]{\scriptsize(+8.0)}\end{tabular} & \begin{tabular}[t]{@{}r@{}}$0.364$\\[-2.5pt]{\scriptsize(+0.4)}\end{tabular} & \begin{tabular}[t]{@{}r@{}}$0.393$\\[-2.5pt]{\scriptsize(+2.7)}\end{tabular} & \begin{tabular}[t]{@{}r@{}}$0.128$\\[-2.5pt]{\scriptsize(-9.7)}\end{tabular} & \begin{tabular}[t]{@{}r@{}}$0.426$\\[-2.5pt]{\scriptsize(-4.7)}\end{tabular} & \begin{tabular}[t]{@{}r@{}}$0.443$\\[-2.5pt]{\scriptsize(-6.3)}\end{tabular} \\
\addlinespace[3pt]
\multicolumn{7}{@{}l}{\emph{full estimator}}\\[1pt]
classifier row & \begin{tabular}[t]{@{}r@{}}$0.156$\\[-2.5pt]{\scriptsize(+0.7)}\end{tabular} & \begin{tabular}[t]{@{}r@{}}$0.348$\\[-2.5pt]{\scriptsize(-11.1)}\end{tabular} & \begin{tabular}[t]{@{}r@{}}$0.373$\\[-2.5pt]{\scriptsize(-11.3)}\end{tabular} & \begin{tabular}[t]{@{}r@{}}$0.130$\\[-2.5pt]{\scriptsize(-9.8)}\end{tabular} & \begin{tabular}[t]{@{}r@{}}$0.396$\\[-2.5pt]{\scriptsize(-21.1)}\end{tabular} & \begin{tabular}[t]{@{}r@{}}$0.414$\\[-2.5pt]{\scriptsize(-22.0)}\end{tabular} \\
endpoint projection & \begin{tabular}[t]{@{}r@{}}$0.151$\\[-2.5pt]{\scriptsize(-2.2)}\end{tabular} & \begin{tabular}[t]{@{}r@{}}$0.360$\\[-2.5pt]{\scriptsize(-3.3)}\end{tabular} & \begin{tabular}[t]{@{}r@{}}$0.384$\\[-2.5pt]{\scriptsize(-4.5)}\end{tabular} & \begin{tabular}[t]{@{}r@{}}$0.144$\\[-2.5pt]{\scriptsize(-0.1)}\end{tabular} & \begin{tabular}[t]{@{}r@{}}$0.420$\\[-2.5pt]{\scriptsize(-10.6)}\end{tabular} & \begin{tabular}[t]{@{}r@{}}$0.439$\\[-2.5pt]{\scriptsize(-10.6)}\end{tabular} \\
path-integrated (\our{}) & $\mathbf{0.154}$ & $\mathbf{0.364}$ & $\mathbf{0.390}$ & $\mathbf{0.144}$ & $\mathbf{0.433}$ & $\mathbf{0.452}$ \\
\bottomrule
\end{tabular}
\end{table}
  \endgroup

\subsection{Post-Processing Hyperparameters}
\label{app:ablation_grid}

  \begingroup
  \renewcommand{\caption}[2][]{\READorigcaption{
    \textbf{Ablation of the channel budget $\kappa$ and cancellation
    exponent $\beta$.}
    Standardised paired differences relative to $\kappa=96$, $\beta=2$;
    positive values favour the variant.
    Bold marks values $\geq 2$ and grey marks values $\leq -2$.
    The reference row is omitted.
    TM, SP, and JO denote temporal, spatial, and joint evaluation.
}}%
  \begin{table}[tbh]
\caption{\textbf{Grid over the channel budget $\kappa$ and the cancellation exponent $\beta$.}
Each cell is the paired sigma of the grid point against the reported configuration
($\kappa=96$, $\beta=2$) on the same $2000$ clips, so \textbf{positive means the point beats
what we ship}; bold marks $\ge+2\sigma$ and grey $\le-2\sigma$. The reference row is omitted
because it is zero by construction; its absolute gaps are those of
\cref{tab:faithfulness}. Levels are temporal (\textsc{tm}), spatial (\textsc{sp}) and joint
(\textsc{jo}). The two protocols disagree systematically: every point that gains under
\textsc{mask} loses under \textsc{prune}. Generated by
\texttt{tools/analysis/tabela\_ablacja.py}.}
\label{tab:ablation_grid}
\centering\small
\setlength{\tabcolsep}{2.6pt}
\begin{tabular}{@{}l rrr rrr rrr rrr@{}}
\toprule
& \multicolumn{6}{c}{Something-Something~v2} & \multicolumn{6}{c}{Kinetics-400} \\
\cmidrule(lr){2-7}\cmidrule(lr){8-13}
& \multicolumn{3}{c}{\textsc{mask}} & \multicolumn{3}{c}{\textsc{prune}}
& \multicolumn{3}{c}{\textsc{mask}} & \multicolumn{3}{c}{\textsc{prune}} \\
\cmidrule(lr){2-4}\cmidrule(lr){5-7}\cmidrule(lr){8-10}\cmidrule(lr){11-13}
Configuration & \textsc{tm} & \textsc{sp} & \textsc{jo} & \textsc{tm} & \textsc{sp} & \textsc{jo}
& \textsc{tm} & \textsc{sp} & \textsc{jo} & \textsc{tm} & \textsc{sp} & \textsc{jo} \\
\midrule
$\kappa=96$, $\beta=0.5$ & $\mathbf{+9.7}$ & $\mathbf{+3.7}$ & $\mathbf{+6.4}$ & $\mathbf{+5.9}$ & \textcolor{gray}{$-2.3$} & \textcolor{gray}{$-4.1$} & $\mathbf{+4.4}$ & $\mathbf{+2.7}$ & $+0.1$ & $+2.0$ & \textcolor{gray}{$-6.6$} & \textcolor{gray}{$-8.6$} \\
$\kappa=96$, $\beta=1$ & $\mathbf{+9.1}$ & $\mathbf{+5.0}$ & $\mathbf{+7.8}$ & $\mathbf{+5.5}$ & $-1.2$ & $-0.8$ & $\mathbf{+5.1}$ & $\mathbf{+5.2}$ & $\mathbf{+2.9}$ & $\mathbf{+3.2}$ & \textcolor{gray}{$-3.1$} & \textcolor{gray}{$-5.9$} \\
$\kappa=128$, $\beta=0.5$ & $\mathbf{+10.1}$ & $\mathbf{+4.2}$ & $\mathbf{+6.1}$ & $\mathbf{+6.3}$ & \textcolor{gray}{$-2.5$} & \textcolor{gray}{$-3.7$} & $\mathbf{+3.9}$ & $\mathbf{+2.9}$ & $+0.7$ & $+1.5$ & \textcolor{gray}{$-5.9$} & \textcolor{gray}{$-8.4$} \\
$\kappa=128$, $\beta=1$ & $\mathbf{+8.7}$ & $\mathbf{+4.5}$ & $\mathbf{+7.0}$ & $\mathbf{+6.1}$ & $-1.2$ & $-1.0$ & $\mathbf{+4.8}$ & $\mathbf{+5.2}$ & $\mathbf{+3.5}$ & $\mathbf{+3.7}$ & \textcolor{gray}{$-2.5$} & \textcolor{gray}{$-4.4$} \\
$\kappa=128$, $\beta=2$ & $-0.3$ & $\mathbf{+3.1}$ & $\mathbf{+3.4}$ & $+0.8$ & $-0.5$ & $+0.1$ & $+1.4$ & $\mathbf{+2.5}$ & $\mathbf{+3.6}$ & $\mathbf{+2.5}$ & $+0.6$ & $-0.3$ \\
$\kappa=256$, $\beta=1$ & $\mathbf{+6.9}$ & $\mathbf{+2.4}$ & $\mathbf{+5.1}$ & $\mathbf{+4.8}$ & \textcolor{gray}{$-3.2$} & \textcolor{gray}{$-2.7$} & $\mathbf{+5.2}$ & $+1.8$ & $+0.6$ & $\mathbf{+4.9}$ & \textcolor{gray}{$-6.6$} & \textcolor{gray}{$-6.9$} \\
$\kappa=256$, $\beta=2$ & \textcolor{gray}{$-2.6$} & $+1.1$ & $+1.8$ & $-1.9$ & \textcolor{gray}{$-2.3$} & $-1.2$ & $+1.3$ & $+0.1$ & $+0.4$ & $\mathbf{+3.4}$ & \textcolor{gray}{$-3.7$} & \textcolor{gray}{$-3.5$} \\
$\kappa=768$, $\beta=1$ & $\mathbf{+6.3}$ & $\mathbf{+2.4}$ & $\mathbf{+5.0}$ & $\mathbf{+3.6}$ & \textcolor{gray}{$-2.9$} & \textcolor{gray}{$-2.9$} & $\mathbf{+4.6}$ & $+0.7$ & $+0.2$ & $\mathbf{+4.7}$ & \textcolor{gray}{$-8.1$} & \textcolor{gray}{$-7.7$} \\
$\kappa=768$, $\beta=2$ & \textcolor{gray}{$-2.9$} & $+1.6$ & $\mathbf{+2.0}$ & \textcolor{gray}{$-2.8$} & \textcolor{gray}{$-2.4$} & $-1.8$ & $+0.9$ & $-0.7$ & $+0.4$ & $\mathbf{+3.4}$ & \textcolor{gray}{$-5.2$} & \textcolor{gray}{$-4.9$} \\
\bottomrule
\end{tabular}
\end{table}
  \endgroup

We examine the channel budget $\kappa$ and the cancellation exponent
$\beta$ within the full estimator.
Since the readout direction determines the channel weights,
we re-evaluate these hyperparameters with the path-integrated
direction in place.
\Cref{tab:ablation_grid} reports standardised paired differences
relative to the reference configuration, $\kappa=96$ and $\beta=2$,
using 2000 clips per dataset under both perturbation protocols.

Reducing $\beta$ generally improves faithfulness under \textsc{mask}.
Under \textsc{prune}, it also improves temporal faithfulness,
but reduces spatial and joint faithfulness.
The effect of cancellation weighting therefore depends on both
the perturbation protocol and the evaluation level.

Increasing the channel budget from 96 to 128 produces relatively
small changes, with modest improvements in several settings.
Larger budgets of 256 and 768 channels provide no consistent
benefit across datasets and protocols, and reduce spatial and joint
faithfulness under \textsc{prune}.
These results support retaining a limited set of channels
for the attribution computation.

We retain $\kappa=96$ and $\beta=2$ as a common configuration
across datasets and evaluation protocols.
The grid shows that alternative settings can improve individual
evaluation scores, but do not provide a consistent improvement
across all settings.

\section{Quantitative Analysis}
\label{app:additional_faithfulness}

\paragraph{Faithfulness protocol.}
We evaluate insertion and most-relevant-first deletion (MoRF)
at 14 fractions, $q = k/16$ for $k=1,\ldots,14$.
Both evaluations use descending attribution rankings at three levels:
temporal, ranking whole frames; spatial, ranking patches within each
frame; and joint, ranking all spatiotemporal cells $(t,i,j)$ together.
Insertion progressively restores the highest-ranked regions,
while deletion progressively removes them.

We use two perturbation protocols.
Under \textsc{mask}, removed regions are replaced by the clip mean,
preserving the sequence length.
Under \textsc{prune}, only the retained tokens enter the scan.
Faithfulness is summarised by
$\mathrm{AUC}_{\mathrm{ins}}-\mathrm{AUC}_{\mathrm{MoRF}}$,
with higher values indicating better performance.
\Cref{fig:curves_insertion,fig:curves_deletion} present insertion
and deletion curves separately to show how each contributes
to the aggregate score.
For a fixed ranking and perturbation protocol, least-relevant-first
deletion retains the same regions as insertion at the complementary
fraction, so we do not report it separately.

\paragraph{Sanity checks.}
We assess whether attribution maps depend on the learned model weights
and the target class. Weight randomisation compares the original map
with one obtained after randomising all model weights.
Class sensitivity compares maps generated for different target classes
on the same input.
The Centre and Random controls are independent of both weights
and target class, so their maps remain unchanged in these tests.
Results are reported in
\cref{tab:localisation_sanity,tab:coverage}.

\subsection{Metric Definitions}
\label{app:metrics}

\paragraph{Spatial localisation.}
We evaluate spatial localisation on 300 Something-Something~v2 clips
with bounding-box annotations from Something-Else, using the same
clips for all methods.
\emph{Pointing} is the fraction of annotated frames whose attribution
maximum falls inside an annotated object box.
\emph{ST-IoU} is the spatiotemporal intersection-over-union between
the top-25\% attribution region and the annotated object tube.
\emph{Energy pointing} is the fraction of positive attribution mass
inside the annotated boxes.
Higher values indicate better localisation for all three metrics.
Results for VideoMamba-S are reported in
\cref{tab:localisation_sanity}, with comparisons across backbone sizes
in \cref{tab:loc_scaling}.

\paragraph{Sanity.}
\emph{Rand.\ wts.} measures the absolute correlation between
the original attribution map and the map obtained after randomising
all model weights; lower values indicate greater sensitivity
to the learned weights.
\emph{Class sens.} in \cref{tab:localisation_sanity} measures
the correlation between maps for the predicted and runner-up classes;
lower values indicate greater differentiation between these targets.
In \cref{tab:coverage}, the comparison instead uses a uniformly
random target class, with correlations closer to zero indicating
less similarity between the maps.

\paragraph{Temporal localisation.}
We evaluate temporal localisation using the annotated procedure steps
in COIN (\cref{tab:temporal_localisation}).
\emph{Peak in step} is the fraction of clips whose temporal relevance
profile reaches its maximum within the annotated step.
\emph{Mass in step} is the fraction of positive temporal relevance
assigned to frames within that step.
\emph{tIoU} measures the temporal intersection-over-union between
the top-25\% highest-ranked frames and the annotated step.
The fraction of sampled frames within the step provides the expected
Peak in step score under uniform random frame selection and
the Mass in step score under uniform positive relevance.

\subsection{Attribution Concentration}
\label{app:concentration}

We examine how attribution mass is distributed across the
spatiotemporal grid using two complementary measures.
\emph{Mass@$\tfrac12$} is the smallest fraction of cells needed
to account for at least half of the total absolute attribution mass,
averaged over clips. Cells are ranked by $|\Phi|$, so lower values
indicate more concentrated maps.
\emph{Contiguity} is the fraction of selected cells that have
at least one selected neighbour, evaluated at the top-$\tfrac{1}{16}$
attribution budget used in the faithfulness evaluation.

As shown in \cref{tab:concentration}, \our{} achieves the lowest
Mass@$\tfrac12$ and the highest Contiguity among the attribution
methods on all three datasets.
Half of its absolute attribution mass is contained in 1.9--2.8\%
of the cells, compared with 6.6--8.6\% for MambaLRP.
These measures describe the distribution of attribution,
but do not establish faithfulness: the fixed Centre control
achieves a Contiguity score of 1.00 despite being independent
of the input, model, and target class.

\begin{table}[t]
\centering
\caption{
    \textbf{Attribution concentration and contiguity across datasets.}
    Bold marks the best value among attribution methods.
    Grey rows denote content-independent controls.
}
\label{tab:concentration}
\setlength{\tabcolsep}{5pt}
\begin{tabular}{@{}l cc cc cc@{}}
\toprule
& \multicolumn{2}{c}{SSv2}
& \multicolumn{2}{c}{K-400}
& \multicolumn{2}{c}{COIN} \\
\cmidrule(lr){2-3}
\cmidrule(lr){4-5}
\cmidrule(lr){6-7}
& Mass@$\tfrac12$ $\downarrow$ & Contig. $\uparrow$
& Mass@$\tfrac12$ $\downarrow$ & Contig. $\uparrow$
& Mass@$\tfrac12$ $\downarrow$ & Contig. $\uparrow$ \\
\midrule
\textcolor{gray}{Centre} & \textcolor{gray}{$0.364$} & \textcolor{gray}{$1.00$} & \textcolor{gray}{$0.364$} & \textcolor{gray}{$1.00$} & \textcolor{gray}{$0.364$} & \textcolor{gray}{$1.00$} \\
\textcolor{gray}{Random} & \textcolor{gray}{$0.293$} & \textcolor{gray}{$0.21$} & \textcolor{gray}{$0.293$} & \textcolor{gray}{$0.21$} & \textcolor{gray}{$0.293$} & \textcolor{gray}{$0.21$} \\
Layer-CAM & $0.176$ & $0.77$ & $0.173$ & $0.72$ & $0.173$ & $0.73$ \\
IntGrad & $0.108$ & $0.51$ & $0.122$ & $0.45$ & $0.113$ & $0.48$ \\
MambaLRP & $0.066$ & $0.87$ & $0.086$ & $0.81$ & $0.071$ & $0.85$ \\
\rowcolor{red!10}
\our{} (ours) & $\mathbf{0.019}$ & $\mathbf{0.90}$ & $\mathbf{0.028}$ & $\mathbf{0.86}$ & $\mathbf{0.026}$ & $\mathbf{0.89}$ \\
\bottomrule
\end{tabular}
\end{table}

\subsection{Further Results}

\paragraph{Faithfulness across backbone sizes.}

\begin{figure}[tbh]
    \centering
    \includegraphics[width=0.86\textwidth]{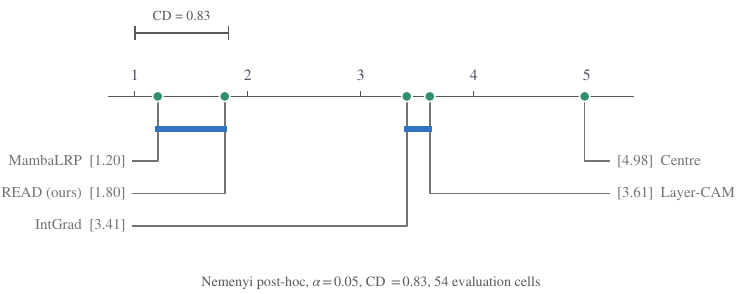}
    \caption{
        \textbf{Average faithfulness ranks across backbone sizes.}
        Results aggregate 54 evaluation settings; lower ranks are better.
        Bars connect methods whose average ranks are not separated
        by the Nemenyi test at $\alpha=0.05$.
    }
    \label{fig:cd_scaling}
\end{figure}

\begin{figure}[tbp]
    \centering
    \begin{subfigure}{\textwidth}
        \centering
        \includegraphics[width=0.85\textwidth]{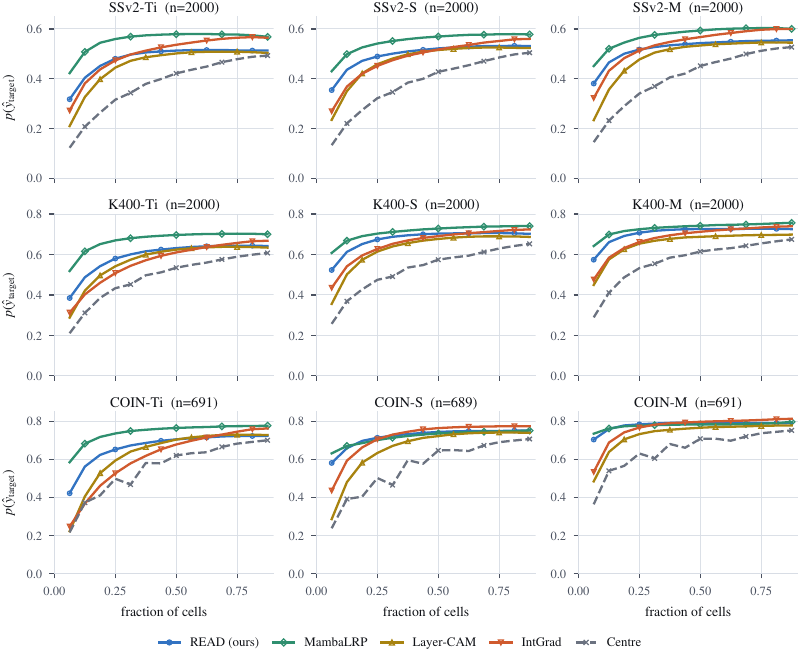}
        \caption{Insertion. Higher values indicate better performance.}
        \label{fig:curves_insertion_scaling}
    \end{subfigure}

    \vspace{1em}

    \begin{subfigure}{\textwidth}
        \centering
        \includegraphics[width=0.85\textwidth]{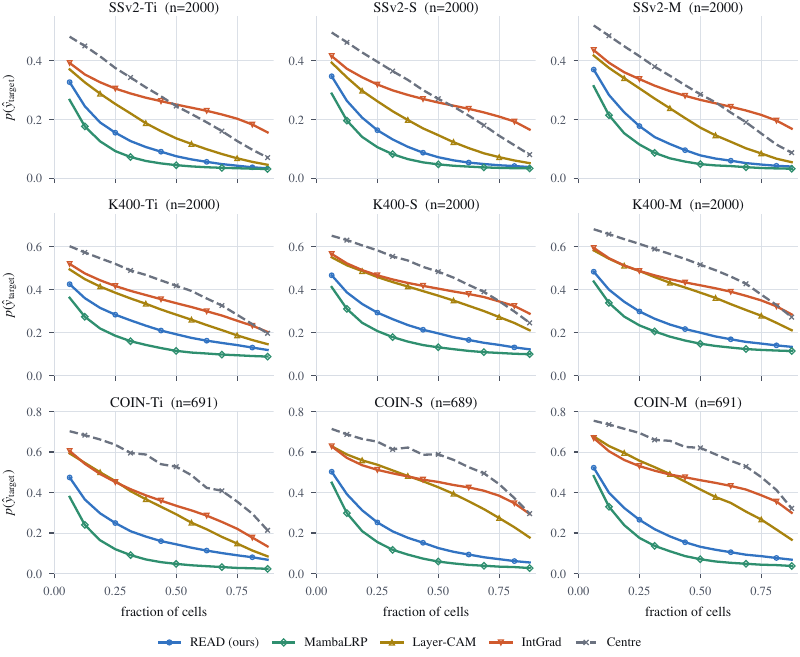}
        \caption{Deletion (MoRF). Lower values indicate better performance.}
        \label{fig:curves_deletion_scaling}
    \end{subfigure}

    \caption{
        \textbf{Joint insertion and deletion curves across datasets and backbone sizes.}
        Results use the \textsc{mask} protocol, with all methods
        evaluated on the same clips within each panel.
    }
    \label{fig:curves_scaling}
\end{figure}

We evaluate faithfulness across all nine dataset--backbone combinations.
\Cref{fig:curves_insertion_scaling,fig:curves_deletion_scaling}
show the joint insertion and deletion curves, while
\cref{fig:cd_scaling} summarises the rankings across evaluation settings.

The overall ordering at the joint level is consistent across backbone
sizes: MambaLRP achieves the highest joint faithfulness scores across
all evaluated backbone--dataset combinations, while the relative
ordering of Layer-CAM and Integrated Gradients varies.
Temporal performance depends more strongly on the backbone.
\our{} outperforms MambaLRP on Kinetics-400 and COIN with the
\textsc{S} and \textsc{M} backbones, but scores lower with \textsc{Ti}.
On Something-Something~v2, MambaLRP achieves higher temporal
faithfulness at all three sizes.

\paragraph{Aggregate rankings.}
\Cref{fig:cd_scaling} summarises average ranks across 54 evaluation
settings: three datasets, three backbone sizes, three attribution
levels, and two perturbation protocols.
MambaLRP and \our{} have the strongest average ranks, which are
not separated by the Nemenyi test at $\alpha=0.05$.
Layer-CAM and Integrated Gradients form a second group.
Because evaluations within each dataset share clips, the settings
are not independent. We therefore use the diagram primarily
to summarise the observed rankings across evaluation settings.

\paragraph{Classification under frame budgets.}
We evaluate whether temporal attribution rankings identify frames
that preserve classification accuracy under a limited frame budget.
For each clip, we retain the $k$ highest-ranked frames and replace
the remaining frames with the clip mean, preserving sequence length
and token order. We then evaluate the frozen classifier's top-1
accuracy against the ground-truth label.
Uniform sampling provides a reference that selects evenly spaced
frames without using attribution scores.

\begin{figure}[tbh]
    \centering
    \includegraphics[width=\textwidth]{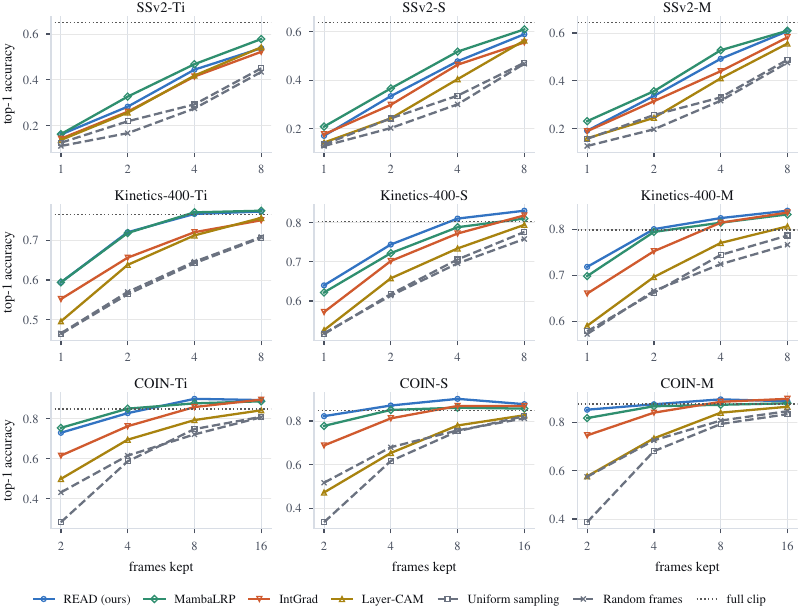}
    \caption{
        \textbf{Classification accuracy under frame budgets.}
        Top-1 accuracy with $k$ selected frames retained and
        the remaining frames replaced by the clip mean.
        Dotted lines indicate full-clip accuracy.
        Uniform sampling selects evenly spaced frames.
    }
    \label{fig:budzet}
\end{figure}

\Cref{fig:budzet} reports results on 491--500 clips per
dataset--backbone combination.
Attribution-based selection improves accuracy over uniform sampling
across the evaluated budgets.
The comparison between \our{} and MambaLRP broadly follows
their temporal faithfulness results:
\our{} performs better on COIN with \textsc{S} and \textsc{M}
and on Kinetics-400 with \textsc{S}, while MambaLRP performs better
on Something-Something~v2 across all three backbone sizes.

On COIN, both methods can exceed full-clip accuracy at moderate
frame budgets. This indicates that masking lower-ranked frames
can improve classification, even though fewer frames remain visible.

\paragraph{Insertion and deletion behaviour.}
\begin{figure}[tbh]
    \centering
    \includegraphics[width=\textwidth]{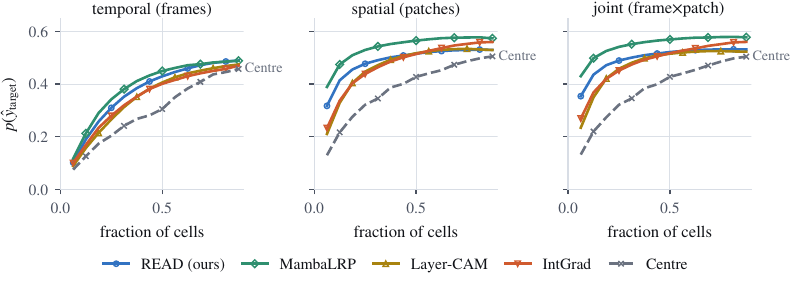}
    \caption{
        \textbf{Insertion curves on Something-Something~v2.}
        Mean target-class probability over 2000 clips under
        the \textsc{mask} protocol, at the temporal, spatial,
        and joint levels. Higher curves indicate better performance.
    }
    \label{fig:curves_insertion}
\end{figure}

\begin{figure}[tbh]
    \centering
    \includegraphics[width=\textwidth]{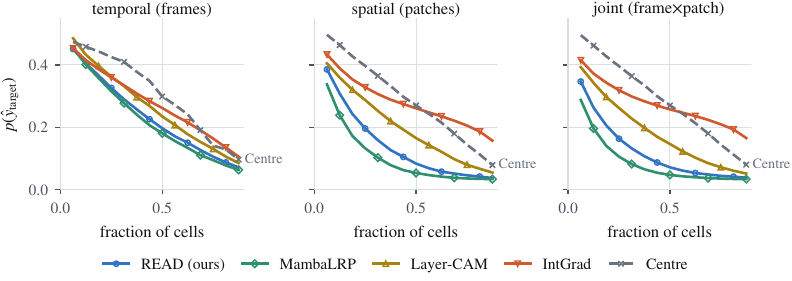}
    \caption{
        \textbf{Deletion (MoRF) curves on Something-Something~v2.}
        The clips, protocol, and evaluation levels match
        \cref{fig:curves_insertion}.
        Lower curves indicate better performance.
    }
    \label{fig:curves_deletion}
\end{figure}

\Cref{fig:curves_insertion,fig:curves_deletion} show the two
components of the faithfulness score on Something-Something~v2.
At the spatial level, \our{} and Integrated Gradients achieve
similar insertion AUCs, but removing regions ranked highly by
\our{} reduces the target-class probability more strongly.
The difference between \our{} and MambaLRP appears in both
components: MambaLRP achieves a higher insertion AUC and
a lower deletion AUC.

\begin{figure}[tbh]
    \centering
    \includegraphics[width=0.86\textwidth]{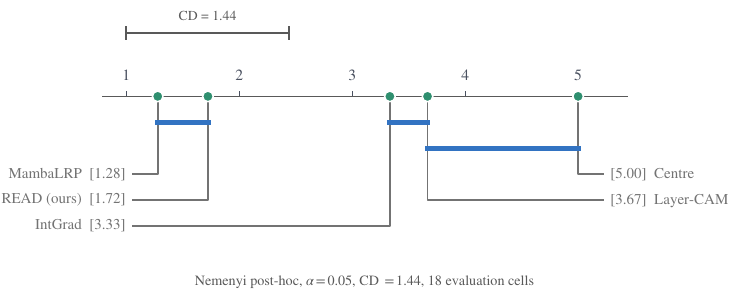}
    \caption{
        \textbf{Average faithfulness ranks for VideoMamba-S.}
        Results aggregate 18 evaluation settings; lower ranks are better.
        Bars connect methods whose average ranks are not separated
        by the Nemenyi test at $\alpha=0.05$.
    }
    \label{fig:cd}
\end{figure}

\paragraph{Rank comparison at VideoMamba-S.}
\Cref{fig:cd} summarises method rankings across 18 evaluation settings:
three datasets, three attribution levels, and two perturbation protocols.
The Friedman test indicates differences across methods, while
the Nemenyi post-hoc test does not separate the average ranks
of \our{} and MambaLRP.

Paired Wilcoxon signed-rank tests with Holm correction detect
differences in clip-level scores, including between these two methods.
These analyses address different questions: the Nemenyi test compares
rankings across evaluation settings, while the Wilcoxon tests assess differences between methods
paired on the same clips.
The dependence between evaluation settings discussed above also
applies to this rank comparison.

On Kinetics-400 with \textsc{Ti} and \textsc{M}, and on COIN
with \textsc{Ti}, the differences between \our{} and MambaLRP
are small relative to the variability of the paired measurements
across all evaluated budgets.
We also measure agreement with the prediction on the unmasked clip.
This produces similar trends to ground-truth accuracy,
so we report only the latter.

\paragraph{Spatial localisation.}
We evaluate spatial localisation on Something-Something~v2 using
the hand and object bounding boxes provided by Something-Else.
The evaluation covers 300 clips and 4645 annotated frames.
Pointing measures how often the attribution maximum falls inside
an annotated box, ST-IoU measures the overlap of the top-25\%
attribution region with the annotated object tube, and Energy
measures the fraction of positive attribution mass inside the boxes.
Formal definitions are provided in \cref{app:metrics}.

  \begingroup
  \renewcommand{\caption}[2][]{\READorigcaption{
    \textbf{Localisation and sanity checks on Something-Something~v2
    with VideoMamba-S.}
    All methods are evaluated on the same 300 clips.
    Higher is better for localisation.
    Sanity checks report the absolute correlation with maps obtained
    after randomising all model weights (\emph{rand.\ wts.}) and
    the correlation between maps for the predicted and runner-up classes
    (\emph{class sens.}); lower is better for both.
    Bold marks the best attribution method in each column, excluding
    the content-independent controls.
    Dashes indicate quantities not measured. 
}}%
  \begin{table}[h]
\centering
\caption{
    \textbf{Localisation and sanity checks on Something-Something~v2
    with VideoMamba-S.}
    All methods are evaluated on the same 300 clips.
    Higher is better for localisation.
    Sanity checks report the absolute correlation with maps obtained
    after randomising all model weights (\emph{rand.\ wts.}) and
    the correlation between maps for the predicted and runner-up classes
    (\emph{class sens.}); lower is better for both.
    Bold marks the best attribution method in each column, excluding
    the content-independent controls.
    Dashes indicate quantities not measured.
}
\label{tab:localisation_sanity}
\begin{tabular}{@{}lccc cc@{}}
\toprule
& \multicolumn{3}{c}{Localisation $\uparrow$}
& \multicolumn{2}{c}{Sanity $\downarrow$} \\
\cmidrule(lr){2-4}
\cmidrule(lr){5-6}
Method
& Pointing
& ST-IoU
& Energy
& Rand. wts.
& Class sens. \\
\midrule
\textcolor{gray}{Centre}
& \textcolor{gray}{$0.70$}
& \textcolor{gray}{$0.32$}
& \,---
& \textcolor{gray}{$1.00$}
& \textcolor{gray}{$1.00$} \\
\textcolor{gray}{Random}
& \textcolor{gray}{$0.44$}
& \textcolor{gray}{$0.19$}
& \textcolor{gray}{$0.49$}
& \textcolor{gray}{$1.00$}
& \textcolor{gray}{$1.00$} \\
Layer-CAM
& $0.66$
& $0.29$
& $0.68$
& $0.06$
& $0.71$ \\
IntGrad
& $0.70$
& $0.24$
& $0.64$
& $\mathbf{0.01}$
& $\mathbf{0.32}$ \\
MambaLRP
& $\mathbf{0.90}$
& $\mathbf{0.35}$
& $0.81$
& $0.05$
& $0.70$ \\
\rowcolor{red!10}
\our{} (ours)
& $0.81$
& $0.33$
& $\mathbf{0.86}$
& $0.06$
& $0.45$ \\
\bottomrule
\end{tabular}
\end{table}
  \endgroup

As shown in \cref{tab:localisation_sanity}, MambaLRP achieves
the highest Pointing and ST-IoU scores, while \our{} achieves
the highest Energy score. Thus, MambaLRP places its attribution
maximum inside the annotated regions more often, while \our{}
assigns a larger fraction of its positive relevance to those regions.
The same pattern holds across backbone sizes
(\cref{tab:loc_scaling}).

The Centre control achieves substantial localisation scores despite
being independent of the input, model, and target class.
This reflects the tendency of annotated objects to appear near
the image centre and motivates evaluating localisation alongside
faithfulness (\cref{tab:faithfulness}).

\paragraph{Sanity checks.}
We assess whether attribution maps depend on the learned model weights
and the target class. Randomising the weights substantially changes
the maps produced by all evaluated attribution methods.
When the target changes from the predicted class to the runner-up,
\our{} produces less correlated maps than MambaLRP and Layer-CAM.
Integrated Gradients achieves the lowest correlation in both checks.
The Centre and Random controls remain unchanged, with correlations
of 1.00, because neither depends on the weights or target class.

Additional sanity and class-specificity results across datasets
and backbone sizes are reported in \cref{tab:coverage}.

\paragraph{Spatial localisation across backbone sizes.}
\Cref{tab:loc_scaling} extends the spatial localisation evaluation
to all three backbone sizes using the same annotated clips.
The ordering is consistent across the evaluated backbones:
MambaLRP achieves the highest Pointing and ST-IoU scores,
while \our{} achieves the highest Energy pointing score.
Thus, the distinction observed for VideoMamba-S also holds
for \textsc{Ti} and \textsc{M}: MambaLRP more often places
the attribution maximum inside the annotated regions,
while \our{} assigns a larger fraction of positive relevance
to those regions.

  \begingroup
  \renewcommand{\caption}[2][]{\READorigcaption{
    \textbf{Spatial localisation across backbone sizes
    on Something-Something~v2.}
    Higher is better for all metrics.
    Bold marks the best attribution method per column,
    excluding the content-independent controls.
}}%
  \begin{table}[tbh]
\caption{\textbf{Spatial localisation across backbone sizes}, Something-Something~v2, the only
dataset with frame-level box annotations. Columns are the three localisation measures of
\cref{tab:localisation_sanity}, each repeated at the three backbone sizes; higher is better and
bold marks the best method in each column among the attribution methods. The controls are kept
because they are informative here rather than formal: \textsc{centre} reaches a pointing score
of $0.70$ without seeing the input or the class, which bounds what localisation alone can
establish. Generated by \texttt{tools/analysis/tabela\_lokskala.py}.}
\label{tab:loc_scaling}
\centering\small
\setlength{\tabcolsep}{4.5pt}
\begin{tabular}{@{}l ccc ccc ccc@{}}
\toprule
& \multicolumn{3}{c}{Pointing $\uparrow$} & \multicolumn{3}{c}{ST-IoU $\uparrow$}
& \multicolumn{3}{c}{Energy $\uparrow$} \\
\cmidrule(lr){2-4}\cmidrule(lr){5-7}\cmidrule(lr){8-10}
Method & \textsc{Ti} & \textsc{S} & \textsc{M} & \textsc{Ti} & \textsc{S} & \textsc{M}
& \textsc{Ti} & \textsc{S} & \textsc{M} \\
\midrule
\textcolor{gray}{Centre} & \textcolor{gray}{$0.70$} & \textcolor{gray}{$0.70$} & \textcolor{gray}{$0.70$} & \textcolor{gray}{$0.32$} & \textcolor{gray}{$0.32$} & \textcolor{gray}{$0.32$} & \textcolor{gray}{\,---} & \textcolor{gray}{\,---} & \textcolor{gray}{\,---} \\
\textcolor{gray}{Random} & \textcolor{gray}{$0.44$} & \textcolor{gray}{$0.44$} & \textcolor{gray}{$0.44$} & \textcolor{gray}{$0.19$} & \textcolor{gray}{$0.19$} & \textcolor{gray}{$0.19$} & \textcolor{gray}{$0.49$} & \textcolor{gray}{$0.49$} & \textcolor{gray}{$0.49$} \\
Layer-CAM & $0.66$ & $0.66$ & $0.64$ & $0.29$ & $0.29$ & $0.28$ & $0.72$ & $0.68$ & $0.67$ \\
IntGrad & $0.74$ & $0.70$ & $0.72$ & $0.24$ & $0.24$ & $0.24$ & $0.67$ & $0.64$ & $0.64$ \\
MambaLRP & $\mathbf{0.89}$ & $\mathbf{0.90}$ & $\mathbf{0.90}$ & $\mathbf{0.36}$ & $\mathbf{0.35}$ & $\mathbf{0.36}$ & $0.81$ & $0.81$ & $0.83$ \\
\rowcolor{red!10}
\our{} (ours) & $0.84$ & $0.81$ & $0.84$ & $0.34$ & $0.33$ & $0.33$ & $\mathbf{0.89}$ & $\mathbf{0.86}$ & $\mathbf{0.88}$ \\
\bottomrule
\end{tabular}
\end{table}
  \endgroup

\paragraph{Sanity and class specificity across backbones.}
\Cref{tab:coverage} reports weight-randomisation and class-specificity
results across all nine dataset--backbone combinations.
\our{} has the lowest weight-randomisation correlation among
the compared methods in seven of the nine settings.
Layer-CAM has the lowest correlation for the \textsc{S} backbone
on Something-Something~v2 and Kinetics-400.

For class specificity, this evaluation compares the target-class
map with a map for a uniformly random class, whereas
\cref{tab:localisation_sanity} uses the runner-up class.
Layer-CAM produces highly correlated maps across these targets,
with correlations of 0.70--0.77.
Both \our{} and MambaLRP produce more distinct maps:
their absolute correlations remain below 0.06 and 0.14,
respectively.
Weight randomisation is evaluated on 120 clips per setting,
and class specificity on 200 clips, with 119 and 197 clips,
respectively, on COIN.
The content-independent controls are omitted from the table
because their maps remain unchanged in both tests.

  \begingroup
  \renewcommand{\caption}[2][]{\READorigcaption{
    \textbf{Sanity and class specificity across datasets and backbones.}
    Bold marks the lowest absolute correlation after weight randomisation
    (\emph{Sanity}) and the correlation closest to zero for
    a uniformly random target class (\emph{Class}) within each row.
}}%
  \begin{table}[tbh]
\caption{\textbf{Sanity and class specificity across datasets and backbones.}
\emph{Sanity} is the absolute correlation between a map and the map obtained after randomising
every weight of the network; the map should collapse, so \emph{lower is better}. \emph{Class} is
the correlation between the map for the target class and the map for a uniformly random class;
here the target is \emph{zero}, and a strongly negative value is not an improvement, so bold
marks the value closest to zero in the row and the smallest value in the sanity group. The two
positional controls (\textsc{centre}, \textsc{random}) see neither the weights nor the class and
must therefore score exactly $1.000$ in both groups; they do so in all $36$ control cells
measured, which is what certifies the measurement rather than any method, and they are omitted
from the table for space. $120$ clips per sanity cell and $200$ per class cell ($119$ and $197$
on COIN, which has fewer clips available; see \cref{app:datasets}). Generated by
\texttt{tools/analysis/tabela\_pokrycia.py}.}
\label{tab:coverage}
\centering\small
\setlength{\tabcolsep}{4.2pt}
\begin{tabular}{@{}ll rrr rrr@{}}
\toprule
& & \multicolumn{3}{c}{sanity $\downarrow$} & \multicolumn{3}{c}{class $\to 0$} \\
\cmidrule(lr){3-5}\cmidrule(lr){6-8}
Dataset & Backbone & Layer-CAM & MambaLRP & \our{} & Layer-CAM & MambaLRP & \our{} \\
\midrule
SSv2 & \textsc{Ti} & $0.039$ & $0.045$ & $\mathbf{0.019}$ & $0.746$ & $-0.068$ & $\mathbf{0.001}$ \\
 & \textsc{S} & $\mathbf{0.041}$ & $0.048$ & $0.055$ & $0.698$ & $-0.086$ & $\mathbf{-0.028}$ \\
 & \textsc{M} & $0.080$ & $0.047$ & $\mathbf{0.027}$ & $0.726$ & $-0.116$ & $\mathbf{0.005}$ \\
\addlinespace[2pt]
Kinetics-400 & \textsc{Ti} & $0.052$ & $0.042$ & $\mathbf{0.014}$ & $0.751$ & $\mathbf{-0.035}$ & $-0.052$ \\
 & \textsc{S} & $\mathbf{0.058}$ & $0.067$ & $0.074$ & $0.720$ & $\mathbf{0.015}$ & $-0.025$ \\
 & \textsc{M} & $0.082$ & $0.085$ & $\mathbf{0.045}$ & $0.738$ & $-0.067$ & $\mathbf{-0.025}$ \\
\addlinespace[2pt]
COIN & \textsc{Ti} & $0.072$ & $0.058$ & $\mathbf{0.023}$ & $0.768$ & $-0.080$ & $\mathbf{-0.051}$ \\
 & \textsc{S} & $0.077$ & $0.081$ & $\mathbf{0.054}$ & $0.741$ & $-0.137$ & $\mathbf{-0.036}$ \\
 & \textsc{M} & $0.091$ & $0.074$ & $\mathbf{0.014}$ & $0.774$ & $-0.084$ & $\mathbf{-0.034}$ \\
\bottomrule
\end{tabular}
\end{table}
  \endgroup

\paragraph{First-frame concentration.}
We measure how often the temporal relevance profile reaches its
maximum at the first frame, using 300 clips from the evaluation split
for each dataset--backbone combination
In VideoMamba's backward scan, the first frame is processed
immediately before the class-token readout.
This proximity may favour early frames independently of their content.

We compare the observed frequency with the uniform reference $1/T$,
where $T$ is the number of frames.
For highlighting in \cref{tab:first_frame}, we use a threshold
of two binomial standard errors above this reference.
All three baselines exceed the threshold in every evaluated setting.
\our{} remains below it on Something-Something~v2 and COIN
across all backbone sizes, while retaining a higher frequency
of first-frame maxima on Kinetics-400.

  \begingroup
  \renewcommand{\caption}[2][]{\READorigcaption{
    \textbf{First-frame concentration across datasets and backbones.}
    Percentage of clips whose temporal relevance peaks at frame~0.
    Chance denotes the uniform reference $1/T$.
    Bold marks values below the threshold defined in the text.
}}%
  \begin{table}[h]
\caption{\textbf{The first-frame artefact across datasets and backbones.} Percentage of clips
whose temporal profile attains its maximum on frame~$0$, over $300$ validation clips per cell.
The readout position is the end of the backward scan, which reaches the tokens of frame~$0$
last, so an estimator that does not account for that proximity inherits it. \emph{Chance} is
$1/T$ and is the only meaningful reference: with $T=32$ frames on COIN, chance alone accounts
for $3.1\%$. Bold marks cells that are \emph{not} raised above chance ($<2\sigma$, binomial).
Every bold cell is \our{}'s: it is not raised above
chance in six of the nine combinations and never exceeds $28\%$, while \emph{all three}
baselines are above chance in \emph{all nine}, from $+3.2\sigma$ to $+68\sigma$. Kinetics-400 is where \our{}
keeps a residual bias, and no cell of that block is bold. Generated by
\texttt{tools/analysis/tabela\_szczyt.py}.}
\label{tab:first_frame}
\centering\small
\begin{tabular}{@{}llccccc@{}}
\toprule
Dataset & Backbone & \textit{chance} & Layer-CAM & IntGrad & MambaLRP & \our{} (ours) \\
\midrule
SSv2           & \textsc{Ti} & 6.2 & 29.0 & 17.7 & 23.3 & \textbf{3.3} \\
               & \textsc{S} & 6.2 & 89.3 & 10.7 & 35.7 & \textbf{6.7} \\
               & \textsc{M} & 6.2 & 89.0 & 18.3 & 28.0 & \textbf{8.3} \\
\addlinespace[2pt]
Kinetics-400   & \textsc{Ti} & 6.2 & 36.7 & 26.0 & 44.3 & 12.3 \\
               & \textsc{S} & 6.2 & 86.7 & 14.7 & 67.0 & 28.0 \\
               & \textsc{M} & 6.2 & 87.3 & 25.3 & 55.0 & 26.3 \\
\addlinespace[2pt]
COIN           & \textsc{Ti} & 3.1 & 16.0 & 10.7 & 7.0 & \textbf{1.0} \\
               & \textsc{S} & 3.1 & 71.7 & 9.0 & 12.7 & \textbf{3.7} \\
               & \textsc{M} & 3.1 & 67.7 & 11.7 & 11.7 & \textbf{1.0} \\
\bottomrule
\end{tabular}
\end{table}
  \endgroup

\section{The Cost of an Explanation}
\label{app:cost}

\begin{figure}[tbh]
    \centering
    \includegraphics[width=\textwidth]{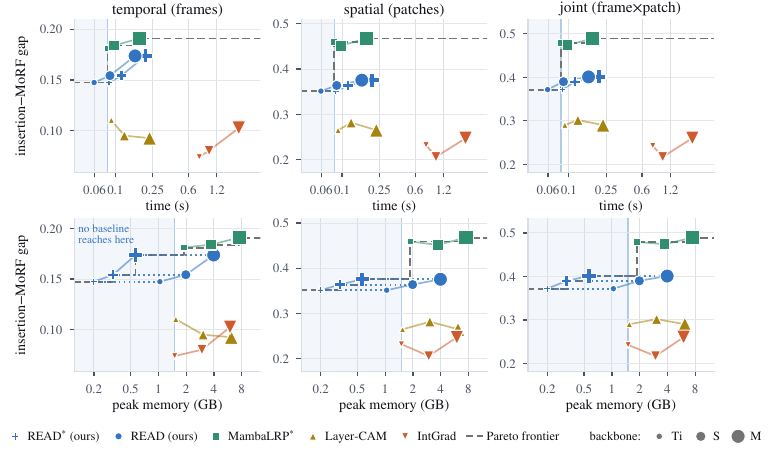}
    \caption{
        \textbf{Faithfulness and computational cost on Something-Something~v2.}
        The insertion--MoRF gap is plotted against runtime (top)
        and peak memory (bottom), with logarithmic cost axes.
        Each point represents a method and backbone pair;
        marker size indicates the backbone.
        Dashed lines show the empirical Pareto frontier.
    }
    \label{fig:cost}
\end{figure}

This section examines the computational cost of attribution across
methods and backbone sizes. \Cref{tab:efficiency_full} reports runtime
and memory usage, including the additional cost relative to a
classification-only forward pass.
\Cref{fig:cost} relates these measurements to faithfulness on
Something-Something~v2.

  \begingroup
  \renewcommand{\caption}[2][]{\READorigcaption{
    \textbf{Computational cost of one attribution map across backbone sizes.}
    Auxiliary costs are relative to a classification-only forward pass.
    Stars denote fused MambaLRP and streaming \our{} implementations.
    Bold marks the lowest cost per backbone among attribution methods.
}}%
  \begin{table}[h]
\caption{
    \textbf{Computational cost of one attribution map across backbone sizes.}
    Auxiliary costs are relative to a classification-only forward pass.
    Stars denote fused MambaLRP and streaming \our{} implementations.
    Bold marks the lowest cost per backbone among attribution methods.
}
\label{tab:efficiency_full}
\centering\small
\setlength{\tabcolsep}{3.6pt}
\begin{tabular}{@{}ll cc rr rr l@{}}
\toprule
& & & & \multicolumn{2}{c}{time (ms) $\downarrow$} & \multicolumn{2}{c}{memory (GB) $\downarrow$} & \\
\cmidrule(lr){5-6}\cmidrule(lr){7-8}
Backbone & Method & Fwd.\ & Bwd.\ & total & aux. & peak & aux. & Auxiliary memory \\
\midrule
\textsc{Ti} & Layer-CAM & 1 & 1 & $91$ & $70$ & $1.55$ & $1.41$ & $\mathcal{O}\!\left(N_L L\right)$ \\
 & IntGrad & 16 & 16 & $781$ & $760$ & $1.50$ & $1.36$ & $\mathcal{O}\!\left(N_L L\right)$ \\
 & MambaLRP & 1 & 1 & $14\,969$ & $14\,947$ & $1.67$ & $1.53$ & $\mathcal{O}\!\left(N_L L\right)$ \\
 & MambaLRP$^{*}$ & 1 & 1 & $83$ & $62$ & $1.88$ & $1.74$ & $\mathcal{O}\!\left(N_L L\right)$ \\
\rowcolor{red!10}
 & \our{} (ours) & \textbf{1} & \textbf{0} & $\mathbf{60}$ & $\mathbf{38}$ & $1.04$ & $0.90$ & $\mathcal{O}\!\left(L\right)$ \\
\rowcolor{red!10}
 & \our{} (ours)$^{*}$ & \textbf{1} & \textbf{0} & $86$ & $64$ & $\mathbf{0.20}$ & $\mathbf{0.06}$ & $\mathcal{O}\!\left(L\right)$ \\
\addlinespace[1pt]
\textit{\scriptsize reference} & \textit{\scriptsize forward pass only} & 1 & 0 & $21$ & --- & $0.14$ & --- & --- \\
\midrule
\textsc{S} & Layer-CAM & 1 & 1 & $125$ & $104$ & $3.06$ & $2.79$ & $\mathcal{O}\!\left(N_L L\right)$ \\
 & IntGrad & 16 & 16 & $994$ & $973$ & $2.97$ & $2.70$ & $\mathcal{O}\!\left(N_L L\right)$ \\
 & MambaLRP & 1 & 1 & $14\,874$ & $14\,853$ & $3.30$ & $3.03$ & $\mathcal{O}\!\left(N_L L\right)$ \\
 & MambaLRP$^{*}$ & 1 & 1 & $97$ & $76$ & $3.73$ & $3.46$ & $\mathcal{O}\!\left(N_L L\right)$ \\
\rowcolor{red!10}
 & \our{} (ours) & \textbf{1} & \textbf{0} & $\mathbf{88}$ & $\mathbf{67}$ & $1.98$ & $1.72$ & $\mathcal{O}\!\left(L\right)$ \\
\rowcolor{red!10}
 & \our{} (ours)$^{*}$ & \textbf{1} & \textbf{0} & $116$ & $95$ & $\mathbf{0.32}$ & $\mathbf{0.05}$ & $\mathcal{O}\!\left(L\right)$ \\
\addlinespace[1pt]
\textit{\scriptsize reference} & \textit{\scriptsize forward pass only} & 1 & 0 & $21$ & --- & $0.27$ & --- & --- \\
\midrule
\textsc{M} & Layer-CAM & 1 & 1 & $231$ & $194$ & $6.17$ & $5.67$ & $\mathcal{O}\!\left(N_L L\right)$ \\
 & IntGrad & 16 & 16 & $2\,054$ & $2\,017$ & $5.98$ & $5.47$ & $\mathcal{O}\!\left(N_L L\right)$ \\
 & MambaLRP & 1 & 1 & $19\,844$ & $19\,807$ & $6.59$ & $6.08$ & $\mathcal{O}\!\left(N_L L\right)$ \\
 & MambaLRP$^{*}$ & 1 & 1 & $180$ & $143$ & $7.52$ & $7.01$ & $\mathcal{O}\!\left(N_L L\right)$ \\
\rowcolor{red!10}
 & \our{} (ours) & \textbf{1} & \textbf{0} & $\mathbf{162}$ & $\mathbf{125}$ & $3.97$ & $3.47$ & $\mathcal{O}\!\left(L\right)$ \\
\rowcolor{red!10}
 & \our{} (ours)$^{*}$ & \textbf{1} & \textbf{0} & $209$ & $172$ & $\mathbf{0.57}$ & $\mathbf{0.06}$ & $\mathcal{O}\!\left(L\right)$ \\
\addlinespace[1pt]
\textit{\scriptsize reference} & \textit{\scriptsize forward pass only} & 1 & 0 & $37$ & --- & $0.51$ & --- & --- \\
\bottomrule
\end{tabular}
\end{table}
  \endgroup

Lines connect backbone sizes within each method.
The Pareto frontier shows the highest measured faithfulness attainable
within a given time or memory budget.
Shaded regions indicate costs below those of every evaluated baseline
configuration. Dotted segments connect \our{} and its streaming
implementation, \our{}$^{*}$, which produce identical attribution maps
and therefore have the same faithfulness scores.

The runtime and memory comparisons reveal different trade-offs.
Both \our{} and fused MambaLRP contribute to the runtime frontier,
while streaming \our{} extends the memory frontier to substantially
lower budgets. Streaming reduces peak memory with some additional
runtime, preserving the quality of the attribution maps.

For MambaLRP, we plot the fused implementation used in our attribution
experiments. The released implementation has substantially higher
runtime but lower peak memory usage; its costs are reported separately
in \cref{tab:efficiency_full}.

\paragraph{Measurement protocol.}
We measure computational cost on a single RTX~4090 with batch size one
and \texttt{fp32}, using five clips and five repetitions per clip
after warm-up. Runtime is measured with CUDA events and explicit
synchronisation. All methods for a given backbone are evaluated
in the same process.

The reference is a classification-only forward pass on the same clip.
Auxiliary time and memory are computed by subtracting the reference
runtime and peak allocation from those measured during attribution.
Total peak memory includes model parameters and inference activations;
the auxiliary columns report the increase over this baseline.

\paragraph{Alternative implementations.}
Starred rows in \cref{tab:efficiency_full} denote implementations
that change computational cost while preserving the attribution
computation.

\emph{MambaLRP$^{*}$} implements the released relevance propagation
rule using the fused selective-scan kernel.
The released implementation processes sequence positions in a Python
loop, which introduces substantial runtime overhead.
When the scan coefficients $\Delta$, $A$, $B$, and $C$ are detached
from the computation graph, the scan is linear in its input.
This allows its propagation step to use automatic differentiation
through the fused kernel. We retain the original treatment of
the pointwise gate and the modified gradient rules outside the scan.

On matched clips, the fused and released implementations produce
numerically equivalent relevance maps, with a maximum relative
error of $1.6 \times 10^{-4}$, matching token rankings and signs,
and the same relevance conservation.
We use the fused implementation for all MambaLRP attribution results.
The released implementation is included in the cost table
as a runtime reference.

\emph{\our{}$^{*}$} accumulates attribution contributions one block
at a time. The token score is a sum of the per-block terms
in \cref{eq:channel_weight}, and the readout direction
in \cref{eq:path_direction} depends only on the current block.
Once its contribution has been accumulated, the block's captured
states can be released.

This streaming computation requires auxiliary memory
$\mathcal{O}(L)$, compared with $\mathcal{O}(N_L L)$ for the evaluated
backward-based implementations, which retain activations across blocks.
These expressions hold other model dimensions fixed.
Streaming introduces additional synchronisation overhead but produces
maps that are bitwise identical to those of the non-streaming
implementation.

\paragraph{Runtime comparison.}
Despite requiring no backward pass, \our{} has a runtime comparable
to fused MambaLRP. The latter uses an optimised selective-scan backward
kernel, while our decomposition combines tensor operations over
the captured states. Runtime therefore depends on both the attribution
computation and its implementation, as well as the hardware.
The timings in \cref{tab:efficiency_full} describe the evaluated
implementations; the memory scaling discussed above follows from
their activation storage requirements.

\paragraph{Runtime breakdown.}
The main computational cost of \our{} comes from evaluating
the per-token contributions to the readout state and projecting them
onto the class-specific direction.
In measurements with otherwise matching implementations,
the path-averaged direction introduces negligible additional runtime
relative to the endpoint projection.
It is computed from the block's readout vectors without additional
model evaluations.
Channel truncation and cancellation weighting
(\cref{sec:composition}) together add approximately 4--8\,ms
across the evaluated backbone sizes.

\paragraph{Memory scaling.}
The memory difference between streaming \our{} and fused MambaLRP
increases with backbone size. Auxiliary memory for fused MambaLRP
rises from 1.74\,GB at \textsc{Ti} to 7.01\,GB at \textsc{M},
while streaming \our{} uses 0.05--0.06\,GB across these configurations
(\cref{tab:efficiency_full}).

The larger backbone also increases depth: \textsc{M} has 32 blocks,
compared with 24 in \textsc{Ti} and \textsc{S}.
Streaming memory depends on sequence length and model width,
but does not accumulate across blocks.

\paragraph{Computational cost.}
\Cref{tab:efficiency_full} summarises attribution time and memory usage
across the three backbone sizes.
Fwd.\ and Bwd.\ denote the numbers of forward and backward evaluations.
Auxiliary memory complexity is expressed in the number of blocks $N_L$
and sequence length $L$, with other model dimensions held fixed.

\clearpage
\section{Extended Qualitative Evaluation}
\label{app:qualitative}

\begin{figure}[h]
    \centering
    \begin{subfigure}[b]{\textwidth}
        \centering
        \includegraphics[
            width=\linewidth,
            trim=0.5cm 8.34cm 3.38cm 0.33cm,
            clip
        ]{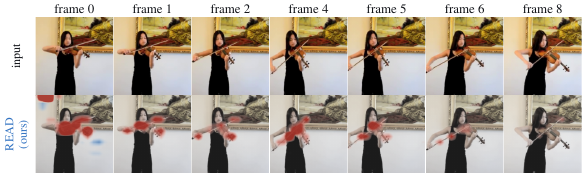}
    \end{subfigure}\hfill
    \begin{subfigure}[b]{\textwidth}
        \centering
        \includegraphics[
            width=\linewidth,
            trim=0.5cm 8.34cm 3.38cm 0.33cm,
            clip
        ]{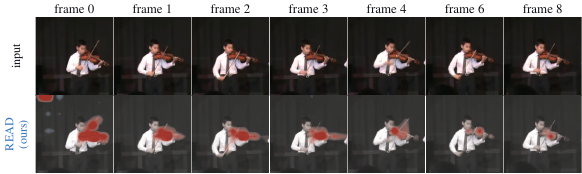}
    \end{subfigure}
    \caption{\textbf{\our{} explanations for two Kinetics-400 clips of
``playing violin''.} Both clips are correctly classified by VideoMamba-S.
Each panel shows seven of the 16 input frames (top) and the corresponding
attribution maps (bottom). Red and blue indicate positive and negative
relevance, respectively, on a colour scale symmetric about zero.}
    \label{fig:violins}
\end{figure}

\Cref{fig:violins} illustrates \our{} on two clips of the same action
in different scenes. The maps highlight regions around the violin and the player's hands, with the spatial distribution of relevance changing across frames as the action unfolds. Although temporal relevance is highest in the first frame in both clips,
it remains present across several frames in the opening part of each clip.

\begin{figure}[thb]
\centering
\includegraphics[width=0.85\textwidth,
trim=0cm 0.05cm 0.75cm 0.34cm, clip]{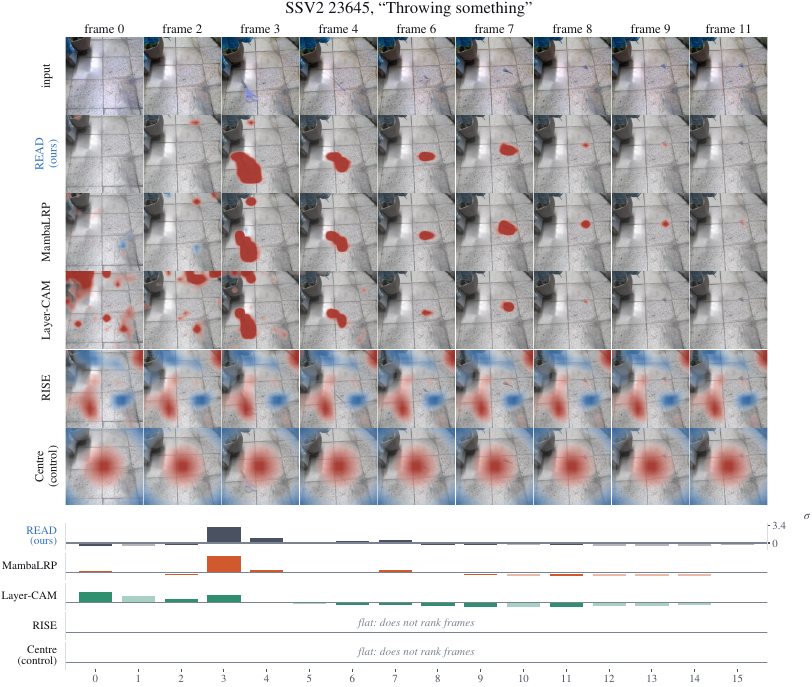}\\
\vspace{3mm}
\includegraphics[width=0.85\textwidth,
trim=0cm 0.05cm 0.75cm 0.34cm, clip]{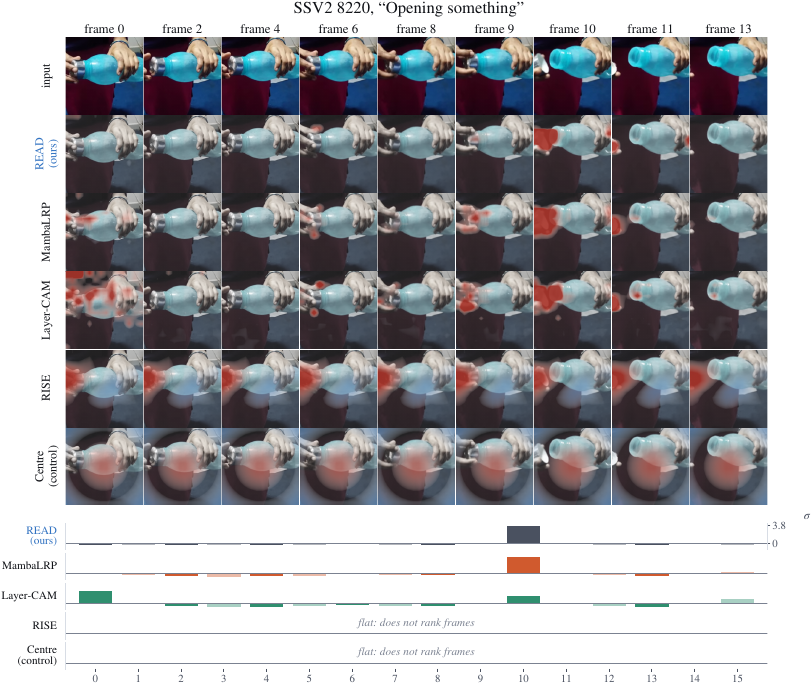}\\
\caption{\textbf{Something-Something~v2.} Classes, top to bottom: \emph{Throwing something},
\emph{Opening something}. Rows within each panel: input, \our{}, MambaLRP, Layer-CAM, RISE and
the static centre control. The bottom strip is the temporal profile $\tau_t$. All maps share one
diverging colormap.}
\label{fig:qualitative_ssv2}
\end{figure}

\begin{figure}[thb]
\centering
\includegraphics[width=0.85\textwidth,
trim=0cm 0.05cm 0.75cm 0.34cm, clip]{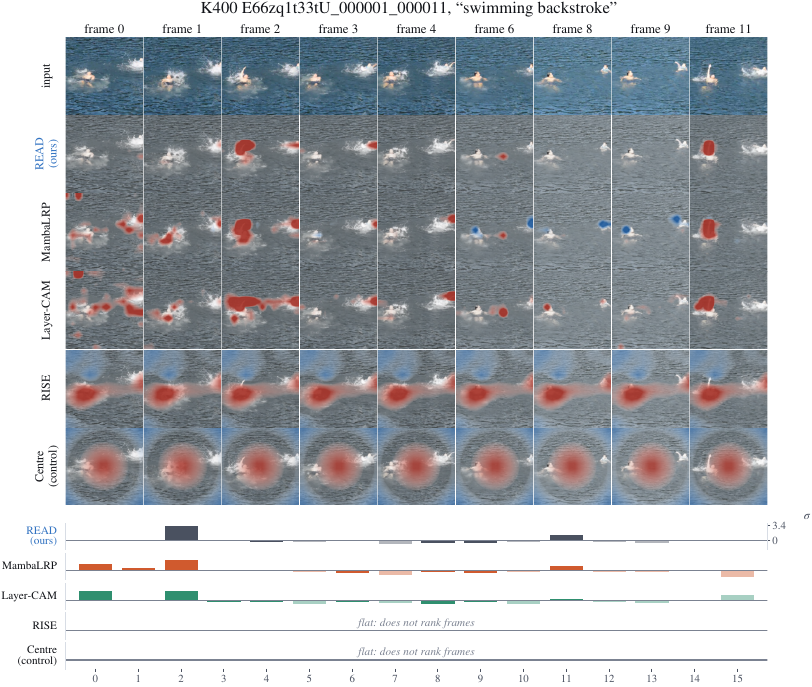}\\
\vspace{3mm}
\includegraphics[width=0.85\textwidth,
trim=0cm 0.05cm 0.75cm 0.34cm, clip]{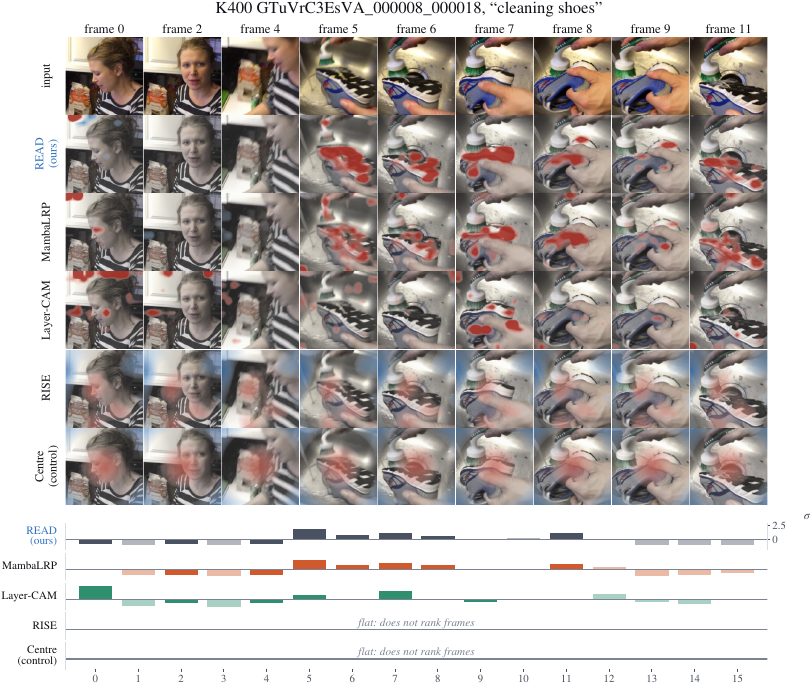}\\
\caption{\textbf{Kinetics-400.} Classes, top to bottom: \emph{swimming backstroke},
\emph{cleaning shoes}. Layout as in \cref{fig:qualitative_ssv2}.}
\label{fig:qualitative_kinetics400}
\end{figure}

\begin{figure}[thb]
\centering
\includegraphics[width=0.85\textwidth,
trim=0cm 0.05cm 0.75cm 0.34cm, clip]{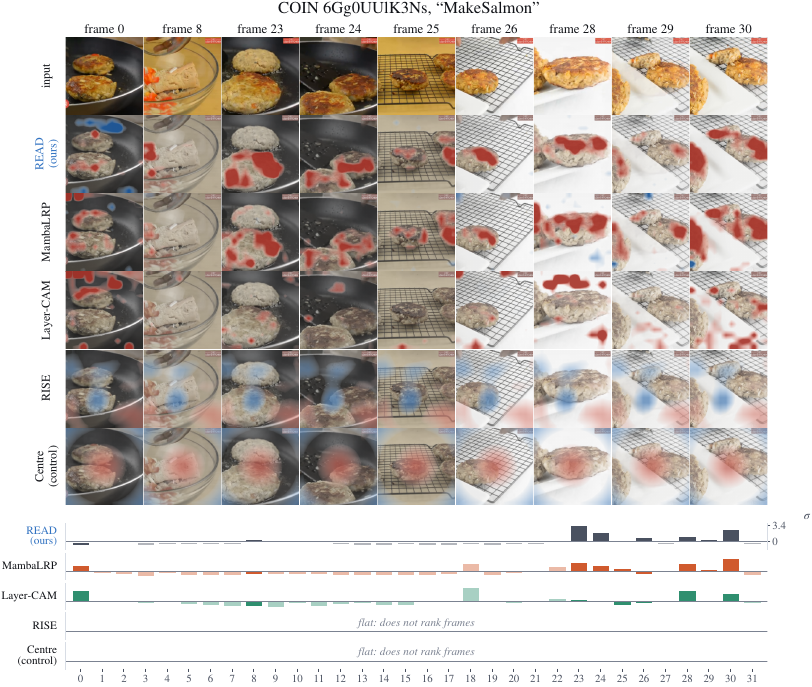}\\
\vspace{3mm}
\includegraphics[width=0.85\textwidth,
trim=0cm 0.05cm 0.75cm 0.34cm, clip]{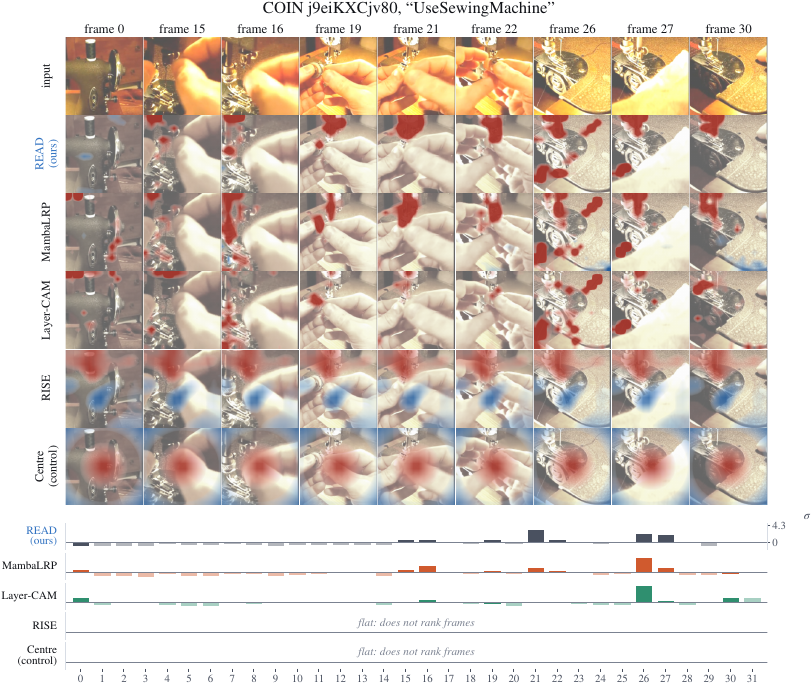}\\
\caption{\textbf{COIN.} Classes, top to bottom: \emph{making a salmon}, \emph{using a sewing
machine}. Layout as in \cref{fig:qualitative_ssv2}; clips here have $32$ frames.}
\label{fig:qualitative_coin}
\end{figure}

\Cref{fig:qualitative_ssv2,fig:qualitative_kinetics400,fig:qualitative_coin}
compare attribution maps for two clips from each dataset.
All methods use the same clips and displayed frames, with temporal
relevance profiles shown below each panel.
The examples were kept fixed across changes to the evaluated configuration.

\paragraph{Perturbation-based comparison.}
We additionally include RISE, which estimates relevance by masking the input
and measuring changes in the model's prediction. Its repeated model
evaluations make attribution substantially more expensive, so we include it
only in the qualitative comparisons. The spatial variant used here applies
each mask across all frames and therefore provides no temporal ranking.
Its temporal profile is consequently flat, as is that of the static
centre control.

\paragraph{Temporal relevance.}
The bar charts show frame-level relevance standardised within each clip,
highlighting how the methods distribute relevance over time.
Layer-CAM frequently assigns the highest relevance to the first frame,
while \our{} also highlights later moments in the action.
These examples complement the aggregate analysis of first-frame
concentration in \cref{tab:first_frame}.

\end{document}